\documentclass{article} %
\usepackage[final]{colm2026_conference}

\usepackage{microtype}
\usepackage{multirow}
\usepackage{hyperref}
\usepackage{fontawesome5}
\usepackage{url}
\usepackage{booktabs}
\usepackage{verbatim}
\usepackage{longtable}

\usepackage[utf8]{inputenc}
\usepackage[T1]{fontenc}
\usepackage{times}
\usepackage{amsmath,amssymb,amsfonts}
\usepackage{pifont}
\usepackage{graphicx}
\usepackage{natbib}
\usepackage{xcolor}
\usepackage{enumitem}
\usepackage{xspace}
\usepackage{tabularx}
\usepackage{adjustbox}
\usepackage{subcaption}
\usepackage{longtable}
\usepackage{titlesec}
\titlespacing{\subsection}{0pt}{5pt}{0pt}
\titlespacing{\section}{0pt}{5pt}{3pt}

\usepackage[capitalize,noabbrev]{cleveref}
\crefname{section}{\S}{\S\S}
\crefformat{section}{\S#2#1#3}
\crefname{figure}{Fig.}{Fig.}
\crefname{algorithm}{Alg.}{Alg.}
\crefname{line}{line}{lines}
\crefname{appendix}{App.}{}
\crefname{equation}{eq.}{eqs.}
\crefformat{equation}{eq.~#2#1#3}
\crefrangeformat{equation}{eqs.~#3#1#4 to #5#2#6}
\crefmultiformat{equation}{eqs.~#2#1#3}{ and~#2#1#3}{, #2#1#3}{ and~#2#1#3}
\crefname{table}{Table}{Tables}
\crefname{proposition}{Proposition}{Propositions}
\crefname{assumption}{Assump.}{Assumps.}
\crefname{definition}{Definition}{Definitions}
\crefname{theorem}{Thm.}{Thm.}

\usepackage[disable]{todonotes}

\newcommand{\tokeval}{\textsc{TokEval}}

\newcommand{\cmark}{\ding{51}}
\newcommand{\xmark}{\ding{55}}

\newif\ifmyflag
\myflagfalse 

\newcommand{\arxivcr}[2]{%
  \ifmyflag
    #1%
  \else
    #2%
  \fi
}

\newcommand{\crnew}[1]{{#1}}

\newcommand{\nPrimary}{29}       %
\newcommand{\nExtended}{39}      %
\newcommand{\nFamilies}{31}      %
\newcommand{\nStdAll}{44}        %
\newcommand{\nStdAllRefs}{46}    %
\newcommand{\nIntrinsicAll}{43}  %
\newcommand{\nMCPrimary}{20}     %
\newcommand{\nMCByteLevel}{18}   %
\newcommand{\nMCExtended}{24}    %
\newcommand{\nMCInStdPanel}{19}  %

\usepackage{lineno}

\definecolor{darkblue}{rgb}{0, 0, 0.5}
\hypersetup{colorlinks=true, citecolor=darkblue, linkcolor=darkblue, urlcolor=darkblue}

\title{TokEval: A Tokenizer Evaluation Suite}

\author{Clara Meister \\
EPFL \\
\texttt{clara.meister@epfl.ch}
}

\begin{document}

\ifcolmsubmission
\fi

\maketitle

\begin{abstract}

Language model tokenizers are typically selected with minimal evaluation, despite the fact that their design choices directly impact model capabilities. This can be partly attributed to a limited understanding of which tokenizer properties affect which aspects of downstream performance. We introduce \tokeval{}, a framework of tokenizer evaluation metrics that goes beyond standard measures like fertility and compression rate to capture linguistically and structurally meaningful properties, e.g., UTF-8 character boundary integrity and digit place-value boundary alignment for mathematics. To validate whether these metrics are predictive of downstream model performance, we conduct controlled language model pretraining experiments, varying solely the tokenizers'  training data mixture, pretokenization strategy, and training algorithm. We evaluate the resulting models on bits-per-byte (a tokenizer-agnostic version of perplexity) and several benchmarks, spanning linguistic understanding, mathematical reasoning, and code generation. Our experiments suggest that different intrinsic properties have different impacts on model abilities: information-theoretic metrics predict language modeling abilities (Spearman $|\rho|$ up to 0.80), while structure-sensitive metrics, such as those measuring digit and line-break handling, correlate with task accuracy. We hope \tokeval{} enables more principled tokenizer evaluation, replacing pretraining sweeps with intrinsic measurement wherever the two agree. 
\vspace{-5pt}
 \begin{center}
     \href{https://github.com/cimeister/tokenizer-intrinsic-evals}{\faGithub~cimeister/tokenizer-intrinsic-evals}\\
     \crnew{\href{https://huggingface.co/cmeister/tokenizer-lm-ablations}{huggingface.co/cmeister/tokenizer-lm-ablations}}
 \end{center}

\end{abstract}

\section{Introduction}

Subword tokenization is a ubiquitous component of modern language model pipelines.
Nearly all contemporary large language models (LLMs) rely on some variant of subword segmentation (e.g., Byte Pair Encoding \citep[BPE;][]{gage1994new,sennrich-etal-2016-neural} or the UnigramLM algorithm \citep{kudo-2018-subword}) to map raw text to discrete token sequences, the format on which models can be efficiently trained and evaluated.
Despite this central role, tokenizer design often receives limited attention. In contrast to the substantial efforts devoted to optimizing model architectures, training data, and learning algorithms, the tokenizer is typically selected using a small set of heuristics, such as vocabulary size, compression rate, or informal inspection of segmentation quality. The choice is then rarely revisited over the course of model development. This neglect persists despite growing evidence that tokenization choices shape downstream model behavior across various domains, including arithmetic accuracy, multilingual performance, and general task quality \citep{ali-etal-2024-tokenizer,singh-strouse-2024-tokenization,liu-etal-2025-superbpe,rust-etal-2021-good,lesci-etal-2025-causal}.
Taken together, these results show that tokenizer design is a substantive modeling choice with measurable consequences for downstream model capabilities.  

Perhaps a core reason for this lack of consideration is that testing out tokenizers is an expensive endeavor: the entire pipeline, from text processing to output formatting, must be reconfigured for each different design choice. 
Furthermore, there are two clear issues with the current intrinsic evaluation protocols:\footnote{Intrinsic evaluation refers to evaluating a tokenizer in isolation, using only the tokenizer and a corpus, without training a downstream model.} 
Firstly, the current set of intrinsic evaluation metrics is very narrow. For example, compression rate tells us only \emph{how much} a tokenizer compresses, but not \emph{how well} it preserves the structural properties that matter for specific downstream tasks.
Second, the relationship between intrinsic metrics and downstream performance remains contested. 
As concrete examples, \citet{zouhar-etal-2023-tokenization} found that R\'{e}nyi efficiency correlates strongly with machine translation quality, and compression rate has long been treated as the default proxy for tokenizer quality \citep{rust-etal-2021-good}.
Subsequent work, however, directly challenged both findings \citep{schmidt-etal-2024-tokenization,cognetta-etal-2024-counterexamples}. 
We argue that these mixed results suggest not that intrinsic evaluation is hopeless, but rather that we need richer intrinsic metrics that capture domain-specific structural tokenizer properties, and controlled experiments, isolating specific impacts of tokenizer characteristics.
 
In this paper, we address both needs. Our contributions are as follows:
\begin{itemize}[nosep,leftmargin=8mm]
     \item We introduce \tokeval{}, an open-source library implementing a suite of intrinsic tokenizer evaluation metrics spanning text, math, code, and multilingual fairness, alongside tokenizer visualizations and integrity checks (\cref{sec:tokeval}).\looseness=-1
    \item We propose novel intrinsic metrics for mathematical content and code that target specific downstream capabilities (\cref{sec:math-metrics}--\cref{sec:code-metrics}).
    \item We conduct controlled pretraining experiments across a grid of tokenizer configurations, varying algorithm, pretokenization strategy, and training data mixture while holding all other variables constant (\cref{sec:results})
    \item We use mixed-effects regression to isolate the contribution of individual metrics while controlling for covariation in the design axes, showing that specific intrinsic metrics are predictive of specific downstream abilities (\cref{sec:results}).
\end{itemize}

\section{Background and Related Work}
\label{sec:related}
 
\paragraph{Tokenization Algorithms.}
To enable a machine learning model to process text, the text must first be represented as a sequence of real-valued vectors. Tokenization is a key step in this preprocessing pipeline. Formally, tokenization maps an input sequence of symbols (typically raw text represented at the byte or character level) to an output sequence of tokens drawn from a finite vocabulary. 
Each token can then be associated with a real-valued vector, for instance, through an embedding lookup table. 
\arxivcr{BPE \citep{gage1994new,sennrich-etal-2016-neural} is the dominant algorithm for learning such a mapping. }{Notably, bytes or characters themselves could be used as the vocabulary symbols, in which case, the mapping is trivial.  However, byte- and character-level tokenization leads to substantially longer input sequences, which incurs significant computational overhead in both training and inference \citep{xue-etal-2022-byt5,tay2022charformer}.
Subword tokenization, where the vocabulary consists of units whose granularity spans between characters and full words, is the dominant approach, as it provides a favorable tradeoff between vocabulary coverage and sequence length. It both provides a useful inductive bias for learning and improves training and inference efficiency. Byte Pair Encoding \citep[BPE; ][]{gage1994new,sennrich-etal-2016-neural}} is the most widely-used subword tokenization algorithm.  
While various other algorithms have been proposed and taken different approaches  \citep[e.g., UnigramLM;][]{kudo-2018-subword}, many have built on BPE, proposing various modifications to address empirically observed issues, e.g., SuperBPE, BoundlessBPE and PickyBPE \citep{liu-etal-2025-superbpe,schmidt2025boundless,chizhov-etal-2024-bpe}.\looseness=-1 
 {
\setlength{\parskip}{0em}

\paragraph{Impacts of Tokenization.}
Several studies have investigated the relationship between tokenization choices and model capabilities. 
\citet{ali-etal-2024-tokenizer} trained language models, ablating only tokenizer settings, and found that tokenizer
choice significantly impacts both LLM downstream performance and training cost. Notably, they saw that standard intrinsic metrics like fertility and parity were not reliably predictive of downstream performance. 
\citet{rust-etal-2021-good} similarly performed tokenizer ablations for LLM training and observed that replacing a multilingual tokenizer with a dedicated monolingual tokenizer improves downstream performance across nearly all tasks and languages, with effect size comparable to that of a language's share of data in the pretraining corpus.  
\citet{singh-strouse-2024-tokenization} demonstrated that digit tokenization directly impacts models' arithmetic capabilities, observing a $\sim20\%$ difference in mathematics task accuracy from changing tokenization strategy alone. 
At a more fundamental level, \citet{lesci-etal-2025-causal} showed \emph{causally} that including a subword in the vocabulary can increase the probability assigned to the corresponding string by up to 17$\times$, compared with representing the string as two tokens. 
\arxivcr{}{\citet{pawar-etal-2025-broken} similarly showed that the splitting of natural words into multiple tokens negatively impacts model performance. }
\arxivcr{}{\citet{chai-etal-2024-tokenization} showed that subword tokenization leaves models sensitive to typos and other text-format perturbations. }
Closest to the model analysis portion of this work, \citet{altintas2025toksuite} release fourteen models pretrained identically except for their off-the-shelf tokenizer, decoupling tokenizer choice from other training factors; they then analyze resulting differences in model behavior. In our model analyse, we instead vary design choices for custom-trained tokenizers and look specifically at how intrinsic metrics  relate to downstream performance. 
Tokenizers have also been linked to discrepancies in model cost and accessibility.
Both \citet{petrov-etal-2023-language} and \citet{ahia-etal-2023-languages} documented systematic cross-lingual unfairness introduced at the tokenization stage, sometimes described as a ``token tax'' on under-represented languages \citep{lundin-etal-2026-token} because of the economic consequences it imparts. 
\arxivcr{}{More recent work has extended this line of inquiry to investigate dialectal and representational disparities \citep{kanjirangat-etal-2025-tokenization}.}\looseness=-1
 
\paragraph{Intrinsic Tokenizer Evaluation.}
Of particular interest to this work is the development of intrinsic tokenizer metrics and studies of how such metrics relate to downstream model performance. 
\arxivcr{}{An \emph{intrinsic} metric evaluates a tokenizer in isolation, without training or querying a downstream model; it is a function of the tokenizer and a text corpus alone. 
By contrast, an \emph{extrinsic} metric measures the effect of a tokenizer on the performance of a model trained with it (e.g., bits-per-byte, task accuracy).
Intrinsic metrics are attractive because they are cheap: evaluating a tokenizer often takes seconds, while training a language model to measure its extrinsic effect can take days or weeks.}
The most widely used intrinsic metrics are compression-based measures, such as corpus token count and fertility. Prior work, however, has shown that compression-based metrics alone are not a reliable objective for tokenizer design  
\arxivcr{ \citep{schmidt-etal-2024-tokenization,cognetta-etal-2024-counterexamples,lester2024training}.}{. For example, \citet{schmidt-etal-2024-tokenization} showed that minimizing corpus token count does not reliably improve downstream performance, and \citet{cognetta-etal-2024-counterexamples} constructed tokenizers whose R\'{e}nyi efficiency can be increased arbitrarily while \emph{degrading} model performance. \citet{lester2024training} showed that models trained over neurally compressed text, produced by arithmetic coding or gzip, underperform subword-tokenized models despite the far higher compression.} 
More recent work has therefore argued for evaluating tokenizers through multiple complementary intrinsic signals rather than through compression alone \citep{lotz-etal-2025-beyond}.
Another line of work studies tokenizer alignment with morphological units. \citet{arnett-bergen-2025-morphscore} introduce MorphScore, which measures the alignment between token boundaries and morpheme boundaries, and show that tokenization quality helps explain some cross-linguistic differences in language-model performance, though morphological boundary alignment itself is not consistently predictive. 
Relatedly, \citet{soler-etal-2024-impact} show that morphology-preserving segmentations tend to yield better contextualized representations than segmentations that split roots and \citet{kanjirangat-etal-2025-tokenization} show that tokenization parity is predictive of downstream performance on dialectal NLP tasks that rely on syntactic and morphological cues.\looseness=-1  

 }

\section{The \tokeval{} Framework}\label{sec:tokeval}\label{sec:framework}
\newcommand{\vocab}{\mathcal{V}}
\newcommand{\token}{t}
\newcommand{\toksequence}{\mathbf{\token}}
\newcommand{\inputtext}{\mathbf{x}}
\tokeval{} computes a suite of intrinsic metrics for a tokenizer over a user-specified text corpus.
The library supports textual inputs on a per-language basis for fine-grained cross-lingual analysis.
Concretely, let $\mathcal{T}$ be a tokenizer and $\mathcal{C} = \{(\inputtext_i, \ell_i)\}_{i=1}^N$ be a corpus
of texts $\inputtext_i$ with language labels $\ell_i$. We use $|\inputtext|_u$ to denote the length of the text $\inputtext$ in unit $u \in \{\texttt{bytes}, \texttt{chars}, \texttt{words}, \texttt{lines}\}$, where \texttt{bytes} is the default.\footnote{The appropriate unit depends on the analysis. Bytes require no language-specific definition of word or character boundaries, but they penalize scripts that have larger per-character UTF-8 encodings (1 byte for Latin, 3-4 for CJK, 4 for emoji). %
Line counts are appropriate for parallel corpora where each line is a translation, making the count constant across languages and isolating the tokenizer from both linguistic conventions and encoding artifacts.}
Every text is encoded to a token sequence $\toksequence_i = \mathcal{T}(\inputtext_i)$, where each token $\token$ comes from a tokenizer's vocabulary $\vocab$.
All metrics are reported per language and aggregated globally. \cref{tab:metric-summary} summarizes the metrics; additional details and design choice justifications are provided in the following subsections.

\tokeval{} is implemented in Python. It natively handles five tokenizer classes: HuggingFace, which loads both the \texttt{tokenizers} (Rust-accelerated) and \texttt{transformers} formats through one class, SentencePiece~\citep{kudo-richardson-2018-sentencepiece}, a pre-tokenized-corpus class for input that is already segmented into tokens, and ScriptBPE/MinGram, which both use the SCRIPT-encoding classes  \citep{land2025bpestaysscriptstructured}; a \texttt{register\_tokenizer\_class} hook lets users register further tokenizer classes. A unified wrapper ensures identical metric computation regardless of backend. \tokeval{} also includes a ``sanity-check'' module that runs sixteen deterministic checks per tokenizer, e.g., byte coverage, combining-mark handling, digit handling, roundtrip fidelity, and vocabulary reachability, and aggregates them into one pass/warn/fail severity per tokenizer. A visualization module generates visualizations of tokenized text and standard plots from a results set, including per-metric bar charts and per-language and faceted breakdowns, across the tokenizers being compared.
\begin{table*}[h!]
\centering
\caption{Summary of \tokeval{} intrinsic metrics. Notation: $p(t)$ is the relative frequency of vocabulary entry $t$ in the input corpus; $n_t$ is the count of all bigrams in the corpus with $t$ in first position; $n_{t,t'}$ is the count of all trigrams with $(t,t')$ in the first two positions; $\mathcal{A}(c)$ is the set of distinct vocabulary entries observed as successors of the context $c$; $\hat{\inputtext} = \mathcal{T}^{-1}(\mathcal{T}(\inputtext))$ is the round-trip reconstruction; $B_{\mathrm{obs}}$/$B_{\mathrm{ideal}}$ are observed/ideal digit boundary sets (see \cref{sec:math-metrics}); $\mathcal{S}$ is the set of special tokens. }
\label{tab:metric-summary}
\small
\adjustbox{max width=\textwidth}{%
\begin{tabular}{@{}p{3.3cm}p{5.5cm}p{6.8cm}@{}}
\toprule
\textbf{Metric} & \textbf{Formula} & \textbf{Interpretation} \\
\midrule
\multicolumn{3}{@{}l}{\textbf{Compression \& Information-Theoretic} (\cref{sec:compression})} \\[2pt]
Compression rate & $\mathrm{CR}_\ell = \sum_{i:\ell_i=\ell} |\inputtext_i|_u \,/\, \sum_{i:\ell_i=\ell} |\toksequence_i|$ & Multiplier by which tokenization compresses. $\uparrow$ = more compression. \\
Unigram entropy & $H(p(\cdot)) = -\sum_t p(t) \log_2 p(t)$ & Uniformity of vocabulary usage. $\uparrow$ = more even. \\
Vocab.\ utilization & $|\{t \in \vocab : p(t) > 0\}| \,/\, |\vocab|$ & Fraction of vocabulary entries the corpus uses at least once. \\
Token length & $\sum_{i}|\inputtext_i|_{\texttt{chars}} \,/\, \sum_i |\toksequence_i|$ & Mean characters per token occurrence. \\[5pt]
Bigram entropy & $\eta(t) = H(p(\cdot|t))\, / \,\log_2 |\mathcal{A}(t)|$; \newline $\bar\eta = \sum_t n_t\eta(t)/\sum_t n_t$ & Conditional entropy of the next token given the previous one, divided by the entropy the context's own observed successor set could carry. $\bar\eta \!\in\! [0,1]$; $\downarrow$ = predictable sequences. \\[15pt]
R\'enyi efficiency & $\bar H_\alpha(p(\cdot)) = H_\alpha(p(\cdot)) / \log_2|\vocab|$, \newline $H_\alpha(p(\cdot)) = \frac{1}{1-\alpha}\log_2(\sum_t p(t)^\alpha)$ & Generalized uniformity; computed for $\alpha \in \{1,2,2.5,3\}$, and reported here for $\alpha=2$. $\uparrow$ = more uniform token usage. \\[20pt]
Avg.\ token rank & $\bar{r}_\ell = \sum_{i:\ell_i=\ell}\sum_j r(t_{i,j}) \,/\, \sum_{i:\ell_i=\ell} |\toksequence_i|$ &Token frequency rank averaged over all token occurrences, where $r$ ranks by frequency in the corpus. $\downarrow$ = few common tokens dominate. \\
\midrule
\multicolumn{3}{@{}l}{\textbf{Linguistic Alignment} (\cref{sec:linguistic})} \\[2pt]
Fertility & $\mathrm{Fert}_\ell = \text{mean}_{i:\ell_i=\ell}\, |\toksequence_i|/|\inputtext_i|_{\texttt{words}}$ & Tokens per word; default $u{=}\texttt{words}$ (mean of ratios). $1.0$ = one token per word. \\
MorphScore & Morpheme boundary P/R/F1 (external library) & Token--morpheme alignment across 70 languages. \\
\midrule
\multicolumn{3}{@{}l}{\textbf{Multilingual Fairness} (\cref{sec:fairness})} \\[2pt]
TFG (Gini) & $\mathrm{TFG} = \sum_{i,j}|c_i - c_j| \,/\, 2n^2\mu$, \newline $c_\ell = \sum_i|\toksequence_i|/\sum_i|\inputtext_i|_u$ & Cross-lingual inequality of tokenization cost. $0$ = perfect parity. $n =$  number of languages; $\mu=$ mean of per-language costs $c_\ell$ \\[10pt]
Vocab.\ util.\ CoV & $u_\ell = |\{\text{used vocab entries in }\ell\}| / |\vocab|$; \newline $\mathrm{CoV} = \sigma(u_\ell)/\mu(u_\ell)$ over languages $\ell$ & Cross-lingual dispersion of per-language vocabulary utilization ($\mu$ = mean, $\sigma$ = sample std). $\downarrow$ = similar utilization across languages. \\
\midrule
\multicolumn{3}{@{}l}{\textbf{Encoding Fidelity} (\cref{sec:fidelity})} \\[2pt]
Exact match & $\mathbb{1}[\hat{\inputtext} = \inputtext]$, averaged over corpus & Strict round-trip reconstruction rate. \\
CER & $\mathrm{Lev}(\inputtext, \hat{\inputtext})\,/\,|\inputtext|_{\texttt{chars}}$ & Character-level Levenshtein distance, normalized; can exceed 1. \\
UTF-8 completeness & $|\{(i,j): \text{bytes}(t_{i,j})\,  \text{is  valid UTF-8 and }\, t_{i,j} \notin \mathcal{S}\}|/ |\{(i,j): t_{i,j} \notin \mathcal{S}\}$ & Fraction of non-special token occurrences that decode to complete characters. Occurrence-weighted in the corpus. \\
Char split rate & $|\{c : \text{multi-byte, split}\}| \,/\, |\{c : \text{multi-byte}\}|$ & Proportion of multi-byte chars whose bytes span multiple tokens. \\
UTF-8 boundary crossing & Frac.\ of non-special token occurrences spanning $>$1 character and leaving one of them incomplete & Frequency of tokens produced by a merge across a character boundary. \\
\midrule
\multicolumn{3}{@{}l}{\textbf{Digit Tokenization} (\cref{sec:math-metrics})} \\[2pt]
Digit boundary $F_1$ & $F_1$ of $B_{\mathrm{obs}}$ vs.\ $B_{\mathrm{ideal}}$ (right-aligned triples) & Place-value alignment of digit span segmentation. \\
Digit split variability & $H_{\mathrm{pattern}} = -\sum_p (n_p/N)\log_2(n_p/N)$ & Entropy of splitting patterns for same-length digit spans. $\downarrow$ = consistent. \\
Operator isolation & Frac.\ of operator occurrences whose covering tokens carry no non-whitespace character from outside the operator span & Whether operators are fused with adjacent operands. Computed separately on prose, code and math; reported here on the math corpus. \\
\midrule
\multicolumn{3}{@{}l}{\textbf{Code Tokenization} (\cref{sec:code-metrics})} \\[2pt]
AST boundary alignment & Frac.\ of AST leaf nodes fully aligned with token boundaries & Per-category (identifier, keyword, operator, literal, delimiter). \\
Ident.\ fragmentation & Frac.\ of identifiers split across $>$1 token & $\uparrow$ = more fragmentation of programmer-defined names. \\
Indentation consistency & $\rho(\text{nesting depth}, \text{whitespace token count})$ & Spearman correlation; $\uparrow$ = more monotonic increase in token count for more whitespace. \\
\bottomrule
\end{tabular}}

 \vspace{-20pt}
\end{table*}

\subsection{Compression and Information-Theoretic Metrics}
\label{sec:compression}

The library's implementation of \textbf{Compression Rate} uses a ratio of sums: total length of the corpus in the chosen base unit over total length in tokens. This is done rather than a mean of per-text ratios, avoiding bias toward short texts.
 \textbf{Unigram Entropy} is the Shannon entropy of the empirical token frequency distribution, as computed with respect to the input corpus. 
\textbf{R\'enyi Efficiency} generalizes this entropy to other entropy orders via a hyperparameter $\alpha$, recovering Shannon entropy when $\alpha=1$~\citep{zouhar-etal-2023-tokenization}.
To capture sequential structure, \textbf{Bigram Entropy} computes the \textit{conditional} entropy of the next token given the previous one, frequency-weighted by how often each bigram-initial token occurs. This metric is largely inspired by the work of \citet{poelman-etal-2025-tokenization}, although it differs in several ways\arxivcr{, documented in the repository.}{.\footnote{This definition deviates from \citet{poelman-etal-2025-tokenization} in several ways: (i) each conditional entropy is normalized by $\log_2$ of the number of distinct successors (next tokens) that the context had in the corpus, i.e., by the maximum entropy achievable by a distribution over that same number of successors;(ii) the aggregation is frequency-weighted rather than an unweighted mean over contexts; (iii) bigram-initial tokens containing punctuation or digits are included; (iv) there is no windowing. The library also computes the reference definition for comparison.}}
Low bigram entropy indicates highly predictable sequences, which can signal over-fragmentation, a common symptom when a tokenizer was not trained on the language being evaluated~\citep{poelman-etal-2025-tokenization}. 
\textbf{Trigram Entropy} uses the same setup, albeit conditioning on bigrams instead of single tokens.
\textbf{Vocabulary Utilization} is the fraction of the total vocabulary entries that occur at least once in the corpus.
\textbf{Token Length} is the mean number of characters per token over all token occurrences in the corpus.
Finally, \textbf{Average Token Rank} assigns each vocabulary entry a rank by descending frequency (rank 1 = most frequent) and reports the mean frequency rank over all token occurrences in a given language~\citep{limisiewicz-etal-2023-tokenization}. A low value indicates that a few common tokens account for most occurrences, suggesting poor vocabulary utilization.

\subsection{Linguistic Alignment Metrics}
\label{sec:linguistic}

\textbf{Fertility} is approximately the inverse of Compression Rate, differing only in aggregation and default choice of $u$. Cross-lingual variation in fertility is a primary indicator of tokenization-induced unfairness~\citep{petrov-etal-2023-language}: fragmenting low-resource languages into many subwords per word degrades downstream performance~\citep{lesci-etal-2025-causal,pawar-etal-2025-broken}.
\textbf{MorphScore}~\citep{arnett-bergen-2025-morphscore, arnett-etal-2025-morphscore-v2} measures token--morpheme boundary alignment via precision, recall, and F1 over gold-standard morphological segmentations, covering 70 languages.
\tokeval{} integrates MorphScore directly, as a submodule of the fork. 
\arxivcr{}{By default, it uses the MorphScore library's recommended settings.}

\subsection{Multilingual Fairness}
\label{sec:fairness}

\textbf{Tokenizer Fairness Gini (TFG)} measures cross-lingual encoding inequality via a Gini coefficient over per-language tokenization costs $c_\ell = \sum_i |\toksequence_i| / \sum_i |x_i|_u$.  
This metric is best computed with $u = \texttt{lines}$ on a parallel corpus, so that differences in $c_\ell$ reflect tokenizer behavior rather than cross-lingual variation in text length or encoding overhead.
Values near 1 indicate severe cross-lingual inequality.
Tokenization parity \citep{petrov-etal-2023-language} compares these same per-language costs pairwise against a reference language; TFG summarizes the same quantities in a single inequality index, so parity is not reported as a separate metric. 
\tokeval{} additionally provides Lorenz curve visualizations.
\textbf{Vocabulary-Utilization CoV} takes the coefficient of variation across per-language vocabulary utilization $u_\ell$, quantifying differences across languages' use of the vocabulary. A lower CoV means the vocabulary is used to a similar degree across languages; a higher CoV means usage concentrates in some languages. It is undefined when fewer than two languages are present.\looseness=-1

\subsection{Encoding Fidelity}
\label{sec:fidelity}

A tokenizer is not necessarily lossless: the round-trip reconstruction $\hat{\inputtext} = \mathcal{T}^{-1}(\mathcal{T}(\inputtext))$ may differ from the original due to Unicode normalization, UNK substitution, or whitespace handling.
\textbf{Reconstruction Exact Match} and, as a graded complement, \textbf{Character Error Rate (CER)} together characterize round-trip fidelity. CER uses the canonical definition of the metric\arxivcr{. }{ and can exceed 1 when $\hat{\inputtext}$ is substantially longer than $\inputtext$ (e.g., byte-fallback expansion). }
High exact-match failure with near-zero CER indicates minor systematic changes (e.g., normalization); high CER indicates fundamental encoding problems. 
A \textbf{UTF-8 Token Completeness} rate $<\!1$ is not necessarily a defect: byte-fallback mechanisms rely on incomplete byte tokens to represent unseen characters. 
\textbf{Character Split Rate} is stratified and reported by byte width (2-, 3-, 4-byte), since CJK characters and emoji are disproportionately affected. 
\textbf{UTF-8 Boundary Crossing Rate} is the fraction of non-special token occurrences whose bytes span more than one character and leave at least one of those characters incomplete. It captures the frequency of tokens produced by a merge across a character boundary, albeit not including (at least one of) the full characters on either side\arxivcr{.}{ of the boundary.} 

\subsection{Digit Tokenization Quality}
\label{sec:math-metrics}

Tokenizer treatment of numbers directly affects arithmetic reasoning~\citep{singh-strouse-2024-tokenization}. Motivated by these findings, we propose metrics that operationalize behaviors shown to benefit models' math processing.
\textbf{Three-digit Boundary Alignment} evaluates whether digit spans are split at positions consistent with right-aligned grouping into triples\arxivcr{.}{: for each contiguous digit span of length $d$, we define \emph{ideal} boundaries at positions $\{d{-}3, d{-}6, \ldots\}$ from the left and compute precision, recall, and $F_1$ against the tokenizer's observed boundaries.}
For spans with $\leq\!3$ digits, both boundary sets are empty; we set $F_1\!=\!1$ by convention iff the tokenizer keeps the span intact (i.e., $|B_{\mathrm{obs}}| = 0$). Results are bucketed by digit length.
\textbf{Digit Split Variability} captures the consistency of segmentation: inconsistent segmentation of same-length digit spans may inhibit learning generalizable arithmetic patterns.
We group digit spans by length, record the splitting pattern of each occurrence (e.g., \texttt{XX|XXX} vs.\ \texttt{X|XXXX}), and compute the Shannon entropy of the resulting pattern distribution.
\textbf{Operator Isolation} measures whether operators are dedicated tokens rather than fused with adjacent operands (e.g., \texttt{+3}).
The operator set is the 24 operators the library scores, in six categories: arithmetic, assignment, logical and bitwise, shift, comparison and ternary.
An occurrence counts as isolated when no token covering it carries a non-whitespace character from outside the operator's own span. 
The metric is computed on three separate corpora, prose, code and math, and reported per domain; we report math results here. 
\arxivcr{}{The library also pools the three into an occurrence-weighted micro-average over operator occurrences.}
\subsection{Code Tokenization Quality}
\label{sec:code-metrics}

Programming languages have deterministic grammars, so syntactic boundaries can be identified without manual annotation.
Parsing source code produces an abstract syntax tree (AST): a tree whose internal nodes
represent grammatical constructs (function definitions, loops, assignments) and whose leaf
nodes represent the atomic syntactic units that appear in the source text. 
We propose novel metrics that evaluate whether tokenizers respect these boundaries.\footnote{All parsing and identification is done using the python tree-sitter library.}
\textbf{AST Boundary Alignment} checks whether each AST leaf node's character span $[s, e)$ coincides with token boundaries.\footnote{A node is \emph{fully aligned} if both $s$ and $e$ fall on token boundaries.}
The library also reports the two one-sided rates separately; the \textbf{End-Alignment Rate} is the fraction of AST leaf nodes whose end offset $e$ falls on a token boundary, irrespective of $s$.
The alignment rate is reported per node category and aggregated across the 15 programming languages we measure: Bash, C, C++, C\#, Go, Java, JavaScript, Lua, PHP, Python, R, Ruby, Rust, Scala, TypeScript.
\textbf{Identifier Fragmentation} is a complement to AST Boundary Alignment. It is the occurrence-weighted fraction of programmer-defined identifiers that require more than one token to encode.
The related \textbf{Tokens per Identifier} reports the mean number of tokens per programmer-defined identifier occurrence rather than the fraction that take more than one. 
\textbf{Indentation Consistency} targets whitespace-sensitive languages (Python and Haskell), computing the Spearman rank correlation between nesting depth and the number of whitespace-only tokens in leading indentation per line. \arxivcr{}{Haskell is absent from all of our code results because its tree-sitter grammar crashes the parser, so the values reported here are Python only.}

\newcommand{\primarypanel}{P\xspace}
\newcommand{\extendedpanel}{E\xspace}
\newcommand{\mathcodepanel}{M\xspace}
\newcommand{\mathcodepanelextended}{Mx\xspace}
\newcommand{\referencepanel}{R\xspace}
\section{Experimental Setup}\label{sec:experiments}\label{sec:experimental-setup}

We train 1.27B-parameter language models under \nStdAllRefs{} tokenizer configurations: \nStdAll{}  tokenizers trained using different configurations, plus two off-the-shelf reference tokenizers (Mistral-Nemo, LLaMA-3). We train two sets of models, differing only in their pretraining data: a natural-language-focused track and math+code-focused track; we call the models of the first set the \emph{natural-language models} and the models of the second the \emph{math+code models} throughout. 
Within each line of experiments, architecture, model training data, and hyperparameters are held constant; the sole experimental variable is the tokenizer.
\arxivcr{}{Every model reported in this paper, the tokenizer each was trained with, and a per-run table of the metrics reported here are released at \url{https://huggingface.co/cmeister/tokenizer-lm-ablations}.}

\subsection{Tokenizer Configurations}
\label{sec:tokenizer-configs}
 {
\setlength{\parskip}{0em}
For the tokenizers that we train, we focus on ablating three axes: algorithm, normalization/pretokenization strategy, and training data composition. Vocabulary size is held constant at $\sim$128K entries.
This paper uses different overlapping subsets of tokenizer--model pairs in different analyses. The subsets are built from \emph{tokenizer families}: a family groups tokenizers that share the same value on all three design axes and differ only in a training hyperparameter or preset of that shared configuration. 
The \textbf{primary panel} (\primarypanel; $n=\nPrimary$) takes one representative per family;\footnote{Grouping near-duplicate configurations into families, and selecting one representative per family for the primary panel, limits how much the correlation analyses rely on tokenizers that differ only in a minor preset. 
\arxivcr{ We discuss these choices more in \cref{app:stat-protocol}.}{Including every near-duplicate as a separate observation would blatantly violate independence assumptions that the correlation results rely on, biasing standard errors downward and inflating apparent significance for any correlation the shared configuration happens to produce. We discuss these choices more in \cref{app:stat-protocol}.} } it is the set on which the aggregate correlations (\cref{tab:aggregate-correlations,tab:aggregate-correlations-fineweb}) and the per-language regressions (\cref{tab:mixed-effects}) are computed.
The \textbf{extended panel} (\extendedpanel; $n=\nExtended$) contains every member of every family; it is used for the family-mean correlations of \cref{app:robustness} and the predictor fits of \cref{app:heldout-reference-prediction}.
The \textbf{math+code panel} (\mathcodepanel; $n=\nMCPrimary$) holds the custom tokenizers with a math+code model: the \nMCInStdPanel{} primary-panel members that have one, plus one English-only-training-data tokenizer kept with a training defect flagged (\cref{app:tok-inclusion}); the math+code columns of \cref{tab:aggregate-correlations} use only the \nMCInStdPanel{} primary-panel members.
The \textbf{reference panel} (\referencepanel; $n=2$) holds the two off-the-shelf tokenizers, which are excluded from every ranking and aggregate statistic, because their training configurations are largely unknown and cannot be controlled for, and serve instead as held-out prediction targets (\cref{app:heldout-reference-prediction})\arxivcr{.}{; the one exception is the cross-scale ranking-stability check of \cref{app:cross-scale-ranking}, which measures how a fixed set of models reorders between training scales and therefore includes them, as its table states.}
\Cref{app:tok-inclusion} gives our full definition of a tokenizer family with examples and the panel definitions are stated concisely in \cref{tab:panel-overview}.

\paragraph{Algorithms.}
We train tokenizers with five algorithms: BPE~\citep{sennrich-etal-2016-neural}, UnigramLM~\citep{kudo-2018-subword}, SuperBPE~\citep{liu-etal-2025-superbpe}, parity-aware BPE \citep{foroutan-meister-et-al-2025-parity-aware-bpe} and MinGram---a minimalist Unigram-style tokenizer trained using an approximation of EM that more closely mimics inference~\citep{land2026mingramminimalistunigramtokenizer}. 
BPE and UnigramLM tokenizers are trained with the HuggingFace \texttt{tokenizers} library; the SuperBPE, parity-aware BPE and MinGram tokenizers are trained using the authors' published codebases. All custom tokenizers are trained on 10\,GB of text. 
\arxivcr{}{For the SuperBPE tokenizers, which use a two-part training strategy, we use either plain BPE or parity-aware BPE for the seed tokenizer,  inheriting either 64K or 90K merges from this tokenizer before extending to 128K. }

\paragraph{Pretokenization and encoding.}
Simply put, pretokenization splits text before a vocabulary learning algorithm is applied based on a deterministic set of rules, determining what spans of text can become a token in the learned vocabulary. 
Consequently, it is a design choice with large influence \citep{arnett2025explaining}. 
The main pretokenization strategies we explore are a minimal punctuation-only baseline, the GPT-4o regex (camelCase splitting, left-aligned 3-digit groups, English contractions), a Claude-inspired regex, and a right-aligned digit variant that groups numbers to place values. 
In ablations, we try targeted variants of these pretokenization strategies. Further details in \cref{app:pretokenizers}. 
The majority of tokenizers we train use byte-level encoding, which is standard practice. We also train several tokenizers using SCRIPT byte encoding~\citep{land2025bpestaysscriptstructured}, which groups characters by Unicode script into blocks before a subword algorithm is applied.

\paragraph{Tokenizer training data compositions.}
The main suite of custom tokenizers are trained on one of three mixtures: English-only, balanced multilingual (35\% English, 30\% multilingual across 30 languages, 15\% math, 15\% code), or code-heavy (50\% English, 50\% code). We sample English training data from FineWeb-Edu~\citep{penedo2024fineweb}, multilingual data from FineWeb2 \citep{penedo2025fineweb2pipelinescale}, math data from FineMath~\citep{finemath}, and code data from StarCoderData~\citep{li2023starcoder}. The multilingual portion spans 30 languages across 11 writing systems (Latin, Cyrillic, Arabic, Devanagari, Bengali, Tamil, Thai, Hangul, CJK, Greek, Hebrew) and 11+ language families. Per-language sampling weights are proportional to estimated character counts in the original FineWeb2 dataset (\cref{app:data}).
Additional tokenizers are trained for ablations, using configurations that weight the 30 multilingual languages equally rather than proportionally, and that restrict the tokenizer's multilingual training data to only a subset of languages. 
More details can be found in \cref{app:tok-data-ablations-configs}.
}

\subsection{Language Model Training}\label{sec:training-details}
All models use the nanochat architecture~\citep{karpathy2025nanochat}, a decoder-only transformer with 24 layers. The model has a total of  1.27B parameters, of which 682M are in transformer weight matrices. 
For the \textit{natural-language-focused models}, we use the same datasets as for tokenizer training, albeit in different proportions. The token budget follows nanochat's compute-optimal scaling rules. Because generative math and code abilities do not surface under the above natural-language-focused mixture at this model scale, we additionally train \textit{math+code models} from scratch on a $\sim$20B-token mixture, 50\% math and 50\% code by text bytes: math text from MegaMath-Web-Pro and code from The Stack v2's educational subset. Due to compute constraints, this second track covers only a subset of all tokenizers trained: the math+code panel (\mathcodepanel) covers \nMCPrimary{} custom tokenizers and 2 off-the-shelf references.\arxivcr{}{\footnote{
\nMCInStdPanel{} of the \nMCPrimary{} custom tokenizers are also members of \primarypanel. The remaining one, \texttt{claude-english-bpe}, is not: the math+code models were trained before it was realized that this tokenizer lacks a full byte alphabet for fallback.  It is included only in the held-out-tokenizer code BPB and MBPP pass@1 fits.}
} \arxivcr{}{Further math+code models outside that panel are used in the appendix analyses (the \nMCExtended{}-member extended math+code roster of \cref{app:mc20b-results}, the byte-alphabet retrains, and the whitespace control of \cref{tab:cross-regime-code-bpb}).}
Full architectural details and hyperparameter values are in \cref{app:architecture}; training data details are in \cref{app:data}.

\subsection{Evaluation}
\label{sec:evaluation}

\paragraph{Downstream benchmarks and perplexity evaluations.}
For the natural-language-focused models, we consider bits-per-byte (BPB) on FLORES+~\citep{nllb-flores24}, over all 215 FLORES+ languages and, separately, restricted to the 31 languages present in our training data.  
We also evaluate on two minimal-pair agreement benchmarks: BLiMP~\citep{warstadt2020blimp} and MultiBLiMP~\citep{jumelet-etal-2026-multiblimp} (English and multilingual, respectively). Both are loglikelihood-scored under the BOS-prefixed convention described in \cref{app:eval-harness}. 
Performance on Belebele (four-option multilingual reading comprehension) was at chance, so we do not report those results.
The math+code models (\cref{sec:training-details}) are evaluated on GSM8K~\citep{cobbe2021gsm8k}, HumanEval~\citep{chen2021codex}, and on MBPP~\citep{austin2021mbpp}, as well as on bits-per-byte held-out StarCoderData across 7 programming languages (Code BPB), using data disjoint from the training mixture. 
Generation tasks prepend the BOS token to the prompt, stop at the model's EOS token, and heal the prompt boundary \citep[as applied by][]{dagan2024getting}; shot counts, scoring, the healing procedure, and the BPB computation are in \cref{app:eval-harness}.

{\setlength{\parskip}{0em}
\paragraph{Intrinsic--downstream correlation.}
We measure the relationship between intrinsic tokenizer properties and downstream model quality via two complementary analyses.
\emph{Aggregate correlation} computes one global intrinsic metric score and one downstream metric score per tokenizer, reporting the Spearman rank correlation between these values. For the intrinsic metrics, this means intrinsic scores are averaged across languages, where applicable. For most downstream metrics, it is already the case that each model receives only a single value, but otherwise (e.g., in the case of FLORES BPB), we likewise average across languages. 
\emph{Mixed-effects regression} fits $D_{T,L} \sim \beta \cdot I_{T,L} + (1 \mid L)$ for each intrinsic metric $I$  and downstream metric $D$, where $T$ indexes tokenizers and $L$ indexes languages.
We reserve this analysis for metrics that provide scores per language.
The random intercept per language accounts for the fact that different languages have different baseline metric scores (e.g., languages' intrinsic difficulties differ, leading to differing baseline BPBs) regardless of tokenizer choice. 
\arxivcr{}{The intrinsic metric $I$ is standardized (zero mean, unit SD across tokenizers, within language); the downstream metric $D$ is not. This makes the magnitude of $\beta$ comparable across metrics. }
In plain terms, $\beta$ reports the change in $D$ associated with a one-SD change in the intrinsic metric, holding language constant. As a concrete example, for $\beta = +0.01$: a tokenizer one SD higher on that metric is associated with FLORES BPB 0.01 higher (worse), within a given language. 
Alongside each $\beta$ we report a standard error: the expected size of the estimate's variation if the experiment were repeated.
\Cref{app:stat-protocol} describes how these standard errors account for the dependence structure of the tokenizer panels.  
We also report the median within-language Spearman $\tilde{\rho}$ across the 31 languages, as this can be more interpretable.
All $p$-values are adjusted for multiple testing via Benjamini--Hochberg (BH) FDR correction.

}

\section{Results}\label{sec:results}
{
\setlength{\parskip}{0em}
The results and following discussion focus on the intrinsic metric--downstream performance relationship. 
The appendix contains material beyond this: full per-tokenizer downstream evaluations and intrinsic-metric tables (\cref{app:additional-results,app:intrinsic-results}), a structural analysis of math+code results (\cref{app:mc20b-results}), and a per-example validation of code structure metrics' ability to predict performance on five external code models (\cref{app:external-per-example}).

\paragraph{Aggregate intrinsic--downstream correlations.}
\cref{tab:aggregate-correlations} shows Spearman correlations between intrinsic metrics and downstream performance; significant, here and below, means BH-adjusted $p_{\text{adj}} < 0.05$. 
Five information-theoretic metrics (\cref{sec:compression}) are significant predictors of FLORES BPB on the trained languages, with $|\rho|$ from $0.49$ (unigram entropy) to $0.80$ (R\'enyi efficiency, the strongest correlation in the table). Of the 14 intrinsic metrics, only digit boundary $F_1$ significantly predicts Code BPB ($\rho = -0.62$), and only AST alignment predicts MBPP ($\rho = 0.61$); we attribute the latter correlation to the vocabulary's line-break handling, which AST alignment partly reflects (\cref{app:mc20b-results}).
Digit boundary $F_1$ is also the only significant BLiMP predictor ($\rho = -0.51$). 
However, a closer look shows that the correlation's sign coincides with an algorithm split rather than a digit-segmentation gradient: the five UnigramLM tokenizers all have digit boundary\footnote{None of the five UnigramLM tokenizers learns a vocabulary with an entry spanning more than one digit, regardless of pretokenizer; \cref{app:pretokenizers}.} $F_1 = 0.483$ and some of the highest BLiMP scores across the panel (0.824 to 0.841). On the other hand, BPE RightAlign's pretokenizer allows vocabulary entries that span 3 digits, and the tokenizer learns such entries, giving it the panel's highest digit boundary $F_1$ ($1.000$), but its BLiMP score is in the panel's mid-range (0.813 to 0.816; \cref{tab:intrinsic-flores,tab:main-results}). Consistent with this, removing a single UnigramLM tokenizer pushes the digit-boundary-$F_1$/BLiMP correlation just above the corrected significance threshold (\cref{app:robustness}).
The largest correlation in \cref{tab:aggregate-correlations} after the R\'enyi-efficiency/FLORES(trained) cell, digit boundary $F_1$ with FLORES(all) BPB ($\rho = -0.67$, significant), reflects the same algorithm split: the five UnigramLM tokenizers have FLORES(all) BPB from $2.666$ to $2.687$ at the shared $F_1 = 0.483$, while the panel's 19 BPE-algorithm rows lie between $2.618$ and $2.657$ (\cref{tab:main-results}). 
Two metrics measured on code and math corpora also correlate with the natural-language models' BPB metrics. Identifier fragmentation, measured on the code corpus, correlates with val BPB at $\rho = 0.53$ and with FLORES(trained) BPB at $0.57$; operator isolation, measured on the math corpus, correlates with val BPB at $0.54$ and with FLORES(all) BPB at $0.55$; all four correlations are significant.
This result is not intuitive. One possible explanation is  that tokenizer training algorithm is a confounder: these intrinsic metrics split the tokenizer training algorithm design axis and tokenizer training algorithm has a strong impact on the downstream metrics. 
As preliminary evidence for this explanation, when restricted to the 19 primary-panel tokenizers that use the plain BPE algorithm, the four correlations fall to $-0.03$ and $0.17$ for identifier fragmentation, and to $0.48$ ($p = 0.04$ before correction) and $0.22$ for operator isolation.
There is a similarly unintuitive result regarding the fairness metrics. PA-BPE NFC GPT-4o has the lowest Gini coefficient in \cref{tab:intrinsic-flores} ($0.017$) and a fertility ($3.99$) near the table's minimum ($3.95$), while its val BPB ($0.7255$) and FLORES(trained) BPB ($1.186$) are among the highest of the BPE tokenizers trained using the balanced mixture \cref{tab:main-results}.
A possible explanation, which we have not tested, is that parity-aware training assigns more vocabulary entries to low-resource languages; under the skewed LM training mixture, those entries occur rarely, so their embeddings receive few gradient updates. At this training budget, encoding equality may not be optimal for achieving lower BPB.

\begin{table*}[t]
\centering

\caption{Aggregate Spearman $\rho$ between intrinsic metrics and downstream
performance. 
Natural language intrinsic metrics are measured on the 31-language training data subset in FLORES+. AST alignment and identifier fragmentation are measured on code data; operator isolation and digit F1 on math data.
\textbf{Bold} indicates significance, with stars marking the (BH-adjusted) thresholds:  $^{*}$\,$p_{\text{adj}}<0.05$, $^{**}$\,$p_{\text{adj}}<0.01$, $^{***}$\,$p_{\text{adj}}<0.001$.}
\label{tab:aggregate-correlations}
\adjustbox{max width=\linewidth}{
\small
\begin{tabular}{lccccccc}
\toprule
\multirow{2}{*}{\textbf{Intrinsic Metric}} & \multirow{2}{*}{\textbf{Val BPB}} & \multicolumn{2}{c}{\textbf{FLORES}} & \multicolumn{2}{c}{\textbf{Math+code}} & \multirow{2}{*}{\textbf{BLiMP}} & \multirow{2}{*}{\textbf{MultiBLiMP}} \\
\cmidrule(lr){3-4} \cmidrule(lr){5-6}
 &  & \textbf{trained} & \textbf{all} & \textbf{Code BPB} & \textbf{MBPP} &  &  \\
\midrule
R\'enyi eff.\ ($\alpha$=2) & $\textbf{-0.57}^{*}$ & $\textbf{-0.80}^{***}$ & $-0.39$ & $-0.49$ & $0.21$ & $-0.32$ & $0.37$ \\
Digit boundary F1 & $-0.42$ & $-0.41$ & $\textbf{-0.67}^{**}$ & $\textbf{-0.62}^{*}$ & $0.21$ & $\textbf{-0.51}^{*}$ & $0.32$ \\
Ident.\ fragmentation & $\textbf{0.53}^{*}$ & $\textbf{0.57}^{*}$ & $0.36$ & $0.41$ & $-0.21$ & $0.37$ & $-0.45$ \\
Operator isolation (math) & $\textbf{0.54}^{*}$ & $0.38$ & $\textbf{0.55}^{*}$ & $0.39$ & $0.27$ & $0.32$ & $-0.25$ \\
UTF-8 boundary crossing & $0.29$ & $\textbf{0.47}^{*}$ & $-0.21$ & $0.45$ & $-0.42$ & $0.27$ & $-0.41$ \\
Trigram entropy & $-0.44$ & $\textbf{-0.66}^{**}$ & $0.12$ & $-0.45$ & $0.10$ & $-0.05$ & $\textbf{0.48}^{*}$ \\
UTF-8 char split & $0.36$ & $\textbf{0.54}^{*}$ & $-0.34$ & $0.24$ & $-0.24$ & $0.09$ & $-0.40$ \\
Compression rate & $-0.32$ & $\textbf{-0.51}^{*}$ & $0.24$ & $-0.12$ & $0.26$ & $-0.14$ & $0.28$ \\
Bigram entropy & $-0.35$ & $\textbf{-0.52}^{*}$ & $0.00$ & $-0.18$ & $0.04$ & $-0.27$ & $0.35$ \\
Vocab utilization & $0.14$ & $-0.07$ & $\textbf{0.61}^{**}$ & $0.56$ & $0.07$ & $0.21$ & $0.02$ \\
AST alignment & $0.31$ & $0.01$ & $0.36$ & $0.16$ & $\textbf{0.61}^{*}$ & $0.10$ & $-0.05$ \\
Fertility & $0.04$ & $0.22$ & $\textbf{-0.57}^{*}$ & $-0.39$ & $0.09$ & $-0.18$ & $-0.08$ \\
Unigram entropy & $-0.27$ & $\textbf{-0.49}^{*}$ & $0.07$ & $-0.02$ & $0.25$ & $-0.22$ & $0.25$ \\
Gini coefficient & $-0.03$ & $0.18$ & $\textbf{-0.58}^{*}$ & $-0.37$ & $0.08$ & $-0.17$ & $-0.03$ \\
\bottomrule
\end{tabular}
}
\end{table*}

\arxivcr{}{
\paragraph{Dependence on the measurement corpus.} 
This section discusses the impact of corpus choice with respect to the two corpora used in each of the intrinsic--extrinsic metric correlations (i.e., the corpus on which intrinsic metrics are measured and on which extrinsic metrics are measured). 
We first ask: are intrinsic metrics robust to corpus choice? 
Repeating the analysis with intrinsic metrics measured on FineWeb-2/-Edu instead of FLORES+ (\cref{tab:aggregate-correlations-fineweb}) changes the significance status of 7 of the 98 aggregate correlations. The other 91 values keep their status and amongst these, only values that are not significant in either table change by more than 0.25.
Five of the seven values that do change significance status are ones for which a) the respective metric depends on the unit $u$ used for measuring the original text's length, and b) that unit differs between the two tables ($u=\texttt{lines}$ on FLORES+, $u=\texttt{bytes}$ on FineWeb).
Three of those five belong to the Gini coefficient: its correlation with FLORES(all) BPB is significant and negative under FLORES+ measurement ($\rho = -0.58$) but positive and not significant under FineWeb measurement ($+0.38$). 
A reversal in the other direction happens as well: Gini becomes a significant predictor of FLORES(trained) BPB ($+0.62$) under FineWeb measurement while it was not under FLORES+ measurement. 
The difference in text length measurement units can explain this: FineWeb is not parallel, so its Gini uses a byte-normalized tokenization-cost basis, which mixes per-script byte-length differences into the inequality measure. 
The line-normalized FLORES+ Gini is arguably the more representative evaluation.
The other two unit-dependent changes are losses of significance: compression rate's correlation with FLORES(trained) BPB ($-0.51$ to $-0.47$) and fertility's correlation with FLORES(all) BPB ($-0.57$ to $-0.37$).
UTF-8 boundary crossing and trigram entropy also become non-significant predictors on FLORES(trained) BPB and MultiBLiMP accuracy, respectively, when the intrinsic metrics are measured on FineWeb data, but these metrics' values do not depend on $u$, so we do not have a good explanation. 
The correlations discussed above also hold when measuring them over the extended-panel and under a leave-one-out setup, with two qualifications, both reported in \cref{app:robustness}. 
We next investigate the impact of the choice of corpus for extrinsic metric computation. A concrete instance of the dependence on this corpus is BPE GPT-4o (highres). Its tokenizer's training data contains English and 5 of the 30 multilingual languages (\cref{app:tok-data-ablations-configs}); the model's training data contains all 30. Its val BPB, $0.7127$, is the lowest of any panel member in \cref{tab:main-results}.
Val BPB is computed on the validation split of the model training mixture, which is a sample whose per-language proportions match those of the language model training data and is therefore English-dominant (\cref{app:data}).
On FLORES(trained) BPB, computed over the 31 trained languages, 25 of which are absent from this tokenizer's training data, the ordering reverses. 
This tokenizer's FLORES(all) BPB, $2.618$, is the lowest of the BPE rows. What a downstream metric measures thus depends on the evaluation population: allocating the vocabulary to the languages with the largest byte shares in the validation mixture lowers val BPB and raises FLORES(trained) BPB at the same time.
\paragraph{Fertility and FLORES(all) BPB.}
One surprising result is that fragmentation is associated with opposite directions of FLORES BPB change depending on which languages' BPB is measured. On the 31 trained languages, more fragmentation is associated with worse BPB: fertility's within-language mixed-effects coefficient on FLORES(trained) BPB is $+0.018$ BPB per SD, significant (\cref{tab:mixed-effects}). On FLORES(all) BPB, computed mostly on the 184 of 215 FLORES+ languages absent from both the tokenizer's and the model's training data, the association reverses: fertility's aggregate correlation with FLORES(all) BPB is $\rho = -0.57$ and the Gini coefficient's is $-0.58$, both significant (\cref{tab:aggregate-correlations}).\footnote{$\beta$ and $\rho$ values are not directly comparable across the two analyses.} In our tokenizer ablations, this pattern  repeats: restricting the tokenizer's training languages from the 30-language balanced mixture to the 6 \texttt{highres} languages lowers FLORES(all) BPB by $0.0232$, and switching the tokenizer's training data from balanced to code-heavy lowers it by $0.0153$; both changes narrow the tokenizer's vocabulary coverage of the FLORES+ languages. BPB is normalized by bytes, so fragmentation is not penalized by construction; the reversal instead reflects a mismatch between the vocabulary and the language being scored. On the untrained languages, a vocabulary specialized toward the trained languages falls back to short, frequent byte-level or near-byte-level fragments, and the model's predictions there are high-entropy but calibrated. A broader multilingual vocabulary instead produces longer, rarer multi-byte tokens on the same text, and the model's conditional distribution over these tokens is fit to the trained languages they resemble, not to the language actually being scored. Cross-entropy penalizes a confident wrong prediction more than a hedged, uncertain one, so the specialized vocabulary's byte-level fallback scores lower (better) BPB even though it encodes the text less efficiently. One could therefore argue that FLORES(all) BPB measures byte-fallback robustness on unseen languages, not multilingual encoding quality on the trained languages.
A similar argument could also be applied for the result on vocabulary utilization, which correlates positively with FLORES(all) BPB ($\rho = 0.61$, significant).
Results on the three SCRIPT-encoding tokenizers are further data points that provide evidence for the byte-fallback interpretation above. MinGram SCRIPT-enc, MinGram SCRIPT-enc (nl-split), and BPE SCRIPT-enc GPT-4o have the highest cross-language standard deviations of per-language FLORES(all) BPB (the second $\sigma$ column of \cref{tab:main-results}). 
The same three tokenizers have the lowest character split rates in \cref{tab:intrinsic-flores}: $0.001$ compared to $0.008$ or more for every other tokenizer. 
SCRIPT encoding replaces byte fallback with script-block units, so text in a language whose script blocks were unseen in training is encoded into rarer units. We attribute the higher per-language BPB variance to this artifact.

\paragraph{Within-language predictors of FLORES BPB.}
When controlling for per-language complexity with a random intercept via mixed-effects regression, the predictiveness of several metrics changes in comparison to aggregate correlations (\cref{tab:mixed-effects}). For per-language FLORES BPB, the information-theoretic metrics remain highly significant predictors, but UTF-8 character split rate and fertility also become significant. %
Fertility in particular illustrates the discrepancy that can arise between the two types of analyses. 
Its within-language coefficient is the largest in magnitude of any metric ($\beta = +0.018$, $\tilde{\rho} = +0.385$), yet its aggregate correlation with FLORES (trained) BPB  in \cref{tab:aggregate-correlations} is not significant. 
The likely explanation is that languages differ substantially in baseline FLORES BPB: Tamil and Bengali have far higher BPB than German or Spanish, regardless of tokenizer. 
Within any given language, higher fertility reliably accompanies worse BPB, but the aggregate correlation pools this within-language relationship together with the between-language baseline differences, which makes the within-language association harder to detect with only $n=\nPrimary{}$ tokenizers.
Repeating this specification with per-language MultiBLiMP accuracy as the outcome, \crnew{five} of the same nine metrics are significant after BH correction, each in the direction consistent with its FLORES BPB effect (a metric associated with higher BPB is associated with lower accuracy).\arxivcr{}{\footnote{Fertility's coefficient is not significant when using clustered standard errors (\cref{app:stat-protocol}).}} 
Of particular interest is that compression rate is not a significant predictor of MultiBLiMP accuracy, consistent with prior findings that compression alone is not a reliable objective for tokenizer quality.
The collective practical implication is that language, and the corpus an intrinsic metric is computed on, need to be controlled for in this kind of analysis.

\begin{table}[t]
\centering
\caption{Relationship between intrinsic tokenizer metrics (FLORES+) and two per-language downstream outcomes (FLORES BPB  and MultiBLiMP accuracy, 31 and 24 languages, respectively), over the paper's primary panel.
$\beta$: coefficient from mixed-effects regression (units: outcome units (BPB or accuracy) per SD of the metric); $\pm$ values are tokenizer-family-clustered standard errors (\cref{app:stat-protocol}).
$\tilde{\rho}$: median within-language Spearman correlation.
\textbf{Bold}:  indicates significance (BH-FDR on the family-clustered $p$-values). $^{{*}}$\,$p_{\text{adj}}<0.05$, $^{{**}}$\,$p_{\text{adj}}<0.01$, $^{{***}}$\,$p_{\text{adj}}<0.001$.}
\label{tab:mixed-effects}
\small
\begin{tabular}{lcccc}
\toprule
\multirow{2}{*}{\textbf{Intrinsic Metric}} & \multicolumn{2}{c}{\textbf{FLORES BPB}} & \multicolumn{2}{c}{\textbf{MultiBLiMP acc.}} \\
\cmidrule(lr){2-3} \cmidrule(lr){4-5}
 & $\beta$ & $\tilde{\rho}$ & $\beta$ & $\tilde{\rho}$ \\
\midrule
Fertility & $\textbf{+0.018}^{***}${\scriptsize$\pm$0.004} & $\textbf{+0.385}$ & $-0.004${\scriptsize$\pm$0.002} & $+0.060$ \\
UTF-8 char split & $\textbf{+0.014}^{***}${\scriptsize$\pm$0.002} & $\textbf{+0.484}$ & $\textbf{-0.005}^{*}${\scriptsize$\pm$0.002} & $\textbf{-0.068}$ \\
Trigram entropy & $\textbf{-0.013}^{***}${\scriptsize$\pm$0.002} & $\textbf{-0.463}$ & $\textbf{+0.003}^{*}${\scriptsize$\pm$0.001} & $\textbf{-0.001}$ \\
Bigram entropy & $\textbf{-0.011}^{***}${\scriptsize$\pm$0.002} & $\textbf{-0.375}$ & $\textbf{+0.003}^{*}${\scriptsize$\pm$0.001} & $\textbf{+0.017}$ \\
Compression rate & $\textbf{-0.010}^{***}${\scriptsize$\pm$0.003} & $\textbf{-0.385}$ & $+0.002${\scriptsize$\pm$0.002} & $-0.063$ \\
R\'enyi eff.\ ($\alpha$=2) & $\textbf{-0.010}^{***}${\scriptsize$\pm$0.001} & $\textbf{-0.378}$ & $\textbf{+0.003}^{*}${\scriptsize$\pm$0.001} & $\textbf{+0.134}$ \\
Unigram entropy & $\textbf{-0.010}^{***}${\scriptsize$\pm$0.002} & $\textbf{-0.326}$ & $+0.002${\scriptsize$\pm$0.001} & $-0.163$ \\
UTF-8 boundary crossing & $\textbf{+0.004}^{*}${\scriptsize$\pm$0.001} & $\textbf{+0.360}$ & $\textbf{-0.005}^{*}${\scriptsize$\pm$0.001} & $\textbf{+0.003}$ \\
Vocab utilization & $-0.002${\scriptsize$\pm$0.004} & $-0.191$ & $-0.001${\scriptsize$\pm$0.002} & $-0.122$ \\
\bottomrule
\end{tabular}
 \end{table}

\paragraph{Predicting a held-out tokenizer.}
Given a tokenizer's intrinsic metrics, can we predict its downstream performance? To test this, we fit intrinsic-metric predictors for each of four downstream metrics (val BPB, FLORES (trained) BPB, code BPB, MBPP pass@1) on the extended panels (\extendedpanel{} and \mathcodepanel{}). 
Cross-validation is leave-one-family-out: each fold holds out one whole family, so no tokenizer is ever predicted by a fit trained on another member of its family (see \cref{app:stat-protocol}). We then use these fits to predict the two off-the-shelf reference tokenizers' models, which were excluded from every fit. Six of the eight resulting predictions (four downstream metrics times two references) fall inside the 90\% prediction interval, the range calibrated on the cross-validation errors to contain 90\% of held-out observations (\cref{app:heldout-reference-prediction}); LLaMA-3 falls outside it on two targets, FLORES (trained) BPB at $+2.10$ standardized errors and code BPB at $-2.01$ (prediction error divided by the cross-validated root-mean-squared error). Rank transfers less well than the point prediction: among the \nExtended{}-tokenizer extended panel plus the inserted reference, the val-BPB fit places LLaMA-3 at predicted rank 1, while its observed val BPB ranks 21st. The practical use of these fits is therefore screening, not selection: a fit predicts a new tokenizer's BPB to within a wide but calibrated interval before any model is trained, which is enough to discard clearly unsuitable configurations, but it does not order close candidates\arxivcr{.}{, so the final choice among them still requires training runs.}

}

\section{Discussion and Conclusion}
\label{sec:discussion}

\paragraph{What intrinsic metrics can and cannot predict.}
Our results show that different intrinsic metrics are significant predictors of different downstream outcomes, with no single metric significant across all of them. 
Information-theoretic and compression-based metrics are reliable for optimizing encoding efficiency: five of them are significant predictors of FLORES(trained) BPB (\cref{tab:aggregate-correlations}), making them useful for narrowing the tokenizer search space before committing to pretraining. They are not a sole proxy for task accuracy \citep{schmidt-etal-2024-tokenization,cognetta-etal-2024-counterexamples}: 
on our limited downstream assessments (BLiMP and MBPP), the only significant task-accuracy predictors in \cref{tab:aggregate-correlations} are structure-sensitive quantities, digit boundary $F_1$ and AST boundary alignment, and even these reflect differences between design-axis groups (algorithm for the BLiMP correlation, line-break handling across pretokenizers for the MBPP correlation; \cref{sec:results,app:newline-mechanism}) rather than gradients within a single design choice. %
\paragraph{How to measure.}
There are three clear takeaways regarding protocol. First, per-language analysis is essential when evaluating over multilingual data: aggregate correlations can mask or even reverse within-language relationships due to baseline differences across languages, as fertility illustrates (\cref{tab:mixed-effects}); the same reasoning motivates tokenizer designs that give each language its own sub-vocabulary.  
Second, the corpus an intrinsic metric is measured on, and its normalization unit, are part of the metric's definition: the Gini coefficient changes sign between line-normalized FLORES+ and byte-normalized FineWeb measurement (\cref{tab:aggregate-correlations-fineweb}). Third, what a downstream metric measures depends on the evaluation population, and this must be accounted for when drawing conclusions about tokenizer impacts: FLORES(all) BPB, computed mostly on languages absent from training, measures byte-fallback robustness of the language models rather than multilingual encoding quality on the trained languages. %

\paragraph{Seed noise and table resolution.}
We retrained five tokenizer configurations three or four times each, changing only the random seed. The seed standard deviation, the run-to-run variation of a score when the same configuration is retrained with a different seed, is $0.00109$ val BPB. 
Among the (custom) tokenizers of \cref{tab:main-results}, the 20 lowest val BPB rows span $0.7116$ to $0.7185$, a spread of $0.0069$, or 6.3 seed standard deviations.
Among the 20 lowest distinct values (23 rows, up to $0.7207$), 19 of the 22 differences between adjacent rows are smaller than one seed standard deviation.
For the math+code models, every one of the 23 differences between adjacent rows in the sorted MBPP column of \cref{tab:mc20b-results} is smaller than the MBPP seed standard deviation ($0.064$), pooled over five configurations retrained with multiple seeds on the math+code mixture.
Adjacent rows of the results tables therefore mostly differ by less than a seed retrain would move a single model. Differences large enough to discard a configuration are visible in the tables; differences between close candidates are within this noise. This is the same conclusion as for the held-out prediction fits above: these results should be used for screening rather than final selection.

 \arxivcr{}{\paragraph{Limitations.}
Main results use a single architecture (nanochat, decoder-only transformer) at 1.27B parameters. Tokenizer--performance relationships may differ at larger scales, where models have greater capacity to compensate for suboptimal tokenization, or under different architectures (e.g., encoder-decoder, state-space models). Byte-level and latent-patch models \citep[e.g.,][]{pagnoni2024byte} are outside the suite's scope for a different reason: they remove the fixed vocabulary the metrics audit, most intrinsic quantities are degenerate at the byte level, and evaluating such a model currently requires training it, which is exactly the cost intrinsic screening avoids. How to evaluate tokenizer-free architectures without training remains open. On the other hand, these models are likely too small for certain abilities to surface, implying our conclusions for these settings may be noisy.
Our tokenizer ablations leave many configurations untested. 
The benchmark suite, though spanning perplexity, linguistic acceptability, math, and code, does not cover important capabilities like long-context reasoning, instruction following, or open-ended generation quality.
Further, with $n = \nPrimary{}$ tokenizers, the aggregate correlation analysis has limited statistical power, especially for detecting moderate effect sizes. Non-significant correlations  should be interpreted as ``not detected'' rather than ``absent.''}

\arxivcr{}{\paragraph{Conclusion.}
We introduced \tokeval{}, an open-source suite of intrinsic tokenizer metrics spanning text, mathematics, code, and multilingual fairness, and validated it with controlled pretraining across tokenizer configurations at two data mixtures. Intrinsic measurement can replace pretraining sweeps where the two agree: information-theoretic metrics screen for encoding efficiency without training a model, and the held-out prediction analysis (\cref{app:heldout-reference-prediction}) shows that such fits bound a new tokenizer's BPB inside a calibrated interval but do not recover its rank among close candidates. Task accuracy is predicted only by structural metrics targeted at the capability in question, and final selection among close candidates still requires training runs. We hope \tokeval{} makes both steps, the intrinsic screening and the targeted structural checks, standard practice in tokenizer development.}

\section*{Acknowledgments}
We thank Tiago Pimentel, Amit Moryossef, and Craig Schmidt for discussions on the metrics presented in this paper, and Tiago Pimentel additionally for detailed feedback on the manuscript. We thank G\"ul Sena Alt{\i}nta\c{s} for recommendations that improved the manuscript. We thank Sander Land for feedback on the design of the \tokeval{} library and for several design improvements to the codebase. The tokenizer language-modeling ablations reported here were carried out as part of Apertus 2 tokenizer development, with compute provided by the Swiss AI Initiative.

\section*{Ethics Statement}

This work is partly motivated by the observation that tokenization choices
disproportionately affect low-resource languages and non-Latin scripts, leading to higher inference costs and degraded performance on already
underserved communities \citep{petrov-etal-2023-language,
ahia-etal-2023-languages}. By providing tools to measure and compare these disparities (e.g., the tokenizer fairness Gini coefficient and per-language fertility), we hope \tokeval{} can help practitioners
identify and mitigate such inequities early in model development. We note, however, that our multilingual coverage (31
languages across 11 scripts) still excludes the vast majority of the world's languages, and our fairness metrics capture only tokenization-level costs, not directly differences in models' cross-lingual abilities.

\bibliography{colm2026_conference}
\bibliographystyle{colm2026_conference}

\appendix

\section{Tokenizer Training and Configuration Details}\label{app:tokenizers}

\subsection{Pretokenization Configurations}
\label{app:pretokenizers}
Before the tokenization algorithm (BPE, UnigramLM, etc.) is applied, raw text is split into coarse segments called \emph{pretokens} using a deterministic rule. The tokenization algorithm then operates independently within each pretoken: merges (in BPE) or segmentation candidates (in UnigramLM) cannot cross pretoken boundaries. Pretokenization thus defines the maximum granularity of the learned vocabulary; a token can be at most one pretoken long. 
We look at several pretokenization strategies in this work, described below. Each is defined as a HuggingFace pretokenizer object. We always use the \texttt{ByteLevel()} pretokenizer, combining this with other pretokenization choices. When a regex is supplied,  the regex splits the raw text first, and byte-level encoding then maps each pretoken's UTF-8 bytes to printable stand-in characters. Learned tokens are therefore byte sequences that never cross a regex split point.} 
\paragraph{Punctuation.} This pretokenizer is defined using the HuggingFace pretokenizer class. It is instantiated as:
\texttt{Sequence([Punctuation(behavior="Isolated"), ByteLevel(use\_regex=True)])}. It isolates punctuation characters, otherwise relying on the byte-level encoder's default regex (GPT-2) for other split points.

\paragraph{GPT-4o.}
CamelCase-aware splitting, left-aligned 3-digit groups (\verb|\p{N}{1,3}|), English contractions (\verb|'s|, \verb|'t|, \verb|'re|, etc.), preceding-punctuation attachment.

\paragraph{Claude.}
Differs from GPT-4o in: whitespace separated by type (spaces, tabs, newlines independently), no punctuation attachment to preceding words, supports both straight and curly apostrophes. We note that this is an approximation of the Claude pretokenization regex, derived using token counts from the Claude API. We do not know definitively the pretokenization choices that Claude uses.
The approximation was derived through the API's token-count endpoint alone using several series of probes: repetition probes, in which a candidate substring is submitted repeated $k$ times (sublinear growth of the token count in $k$ means the repetitions merge inside one pretokenizer chunk), and additivity probes, in which the count of a concatenation is compared against the sum of the parts' separate counts (a saving means the parts share a chunk). Padding text isolates each probe from the surrounding context. A regex clause was accepted only when both probe types agreed and the result reproduced across several Claude models. 

\paragraph{Right-aligned digits.} Pretokenizes digit spans to groups of three according to place values (e.g., ``1234567'' $\to$ [``1'', ``234'', ``567'']). 
Identical to GPT-4o except the digit group regex uses a lookahead for place-value alignment: \verb|\p{N}{1,3}(?=(?:\p{N}{3})*(?:\P{N}|\$))|. This ensures ``123456'' tokenizes as [``123'', ``456''] (thousands-aligned) rather than left-aligned [``12'', ``345'', ``6''].
Under UnigramLM training, this regex left the learned vocabulary's digit handling unchanged. Neither this variant nor the one using the GPT-4o-regex has a vocabulary entry containing two or more consecutive decimal digits, even though the pretokenizer allows it. In both the only decimal-digit entries are the ten single digits.
The two tokenizers produce identical token-string segmentations on a test set covering digit runs up to 15 digits, comma-grouped numbers, and English, Russian, Tamil, and Japanese text.
The two vocabularies are otherwise near-identical: 11 of the 128{,}256 entries of each are absent from the other.
The panel's other three UnigramLM tokenizers also have no vocabulary entry with two or more consecutive decimal digits.

\paragraph{Clean, plus2, plus3.}
\texttt{Clean} is a fifth, multilingual-oriented regex, distinct from the four above: it does not split camelCase, does not explicitly attach contractions, and treats each digit as its own pretoken rather than grouping digits into runs. \texttt{plus2} extends \texttt{clean} by allowing leading apostrophes (ASCII or curly) to attach to the start of a pretoken, targeting French/Italian/Catalan contractions (e.g., \emph{l'arbre}); \texttt{plus3} additionally attaches a trailing apostrophe not followed by a letter, targeting Maltese \emph{ta'}. 

\subsection{Training Data: Ablation Configurations}\label{app:tok-data-ablations-configs}
The intermediate-language-coverage ablations restrict the tokenizer's training data to 5 of the 30 multilingual languages plus English (\texttt{highres}: rus, spa, deu, fra, cmn, plus English; 6 languages total) or 20 of the 30 plus English (\texttt{highmid}: those same 5 \texttt{highres} languages plus 15 mid-resource languages, dropping the 10 lowest-resource languages, plus English; 21 languages total), so the language model still sees all 30 languages during training but the tokenizer never saw the omitted ones. This is meant to test impact on models' generalization to languages absent from tokenizer training. We also train a set of 5 tokenizers that weight the 30 languages equally rather than proportionally, one of which omits the repeat-sampling the other four use to fill low-resource tails. %

\subsection{Tokenizer Panel Specification}\label{app:tok-inclusion}
This section defines eligibility for the primary and extended panels, the family grouping used to partition them, and the handling of tokenizers excluded from both. \cref{tab:panel-overview} gives an overview of the panel definitions and \cref{tab:tokenizer-grid} specifies all tokenizers. 
 
\paragraph{Eligibility.} Membership in the primary correlation panel (\primarypanel) is criteria-defined. A tokenizer is eligible if it meets all four of the following conditions:
\begin{itemize}
    \item it is trained on the paper's standard source corpus; any composition, subset, or reweighting of that corpus is a value on the training-data axis, so the language-coverage and reweighting ablations remain eligible;
    \item its vocabulary falls in the $\sim$128K size band;
    \item it carries no known tokenizer-training defect;
    \item it is not a deliberate minimal-structure or null-control probe.
\end{itemize}
 
\paragraph{Tokenizer families.} Two tokenizers belong to one family when they share the same value on all three design axes, algorithm, normalization/pretokenization strategy, and training data composition, and differ only in a training hyperparameter or preset of that shared configuration. For example, BPE GPT-4o (all-eq.) and its no-repeat-sampling variant differ only in a sampling option and form one family, whereas BPE GPT-4o (bal.) and BPE GPT-4o NFC differ in a value on the normalization axis and form two families. Options internal to one algorithm's implementation also count as presets: the two MinGram tokenizers, which differ only in whether merges may cross line breaks, form one family under this clause. Treating line-break crossing as a value on the pretokenization axis would instead place them in two families. The \nFamilies{} families partition the \nExtended-member extended panel, and the Fam.\ column of \cref{tab:tokenizer-grid} lists every row's family.
 
\paragraph{Panel construction.} Every member of every family is a member of the \textbf{extended} panel ($n=\nExtended$). The \textbf{primary} panel takes one representative per family, chosen at the family's default preset, except two families, which contribute none.\footnote{Both are SuperBPE configurations. SuperBPE trains in two stages: stage 1 trains a tokenizer for a fixed number of merges (we refer to this as the seed tokenizer; in our runs the seed is plain BPE or parity-aware BPE, \cref{sec:tokenizer-configs}}). Stage 2 extends the seed tokenizer with additional merges under a second pretokenization regex. The stage-2 regex is the regex that segments text when the finished tokenizer is applied, so it is the value the pretokenization axis considers for a SuperBPE configuration; the choice of seed tokenizer, including the regex the seed was trained under, is a training choice we consider internal to the algorithm, on the same footing as a hyperparameter. Each of the two stage-2 regexes with a primary-panel SuperBPE build is represented once, at its default preset. %
Ablations of other design axes remain their own families in the extended-level analyses, which operate on them as units. This gives the primary panel \nPrimary{} members. The primary panel's members are therefore near-independent design points by construction; strong dependencies between points would violate independence assumptions important to the validity of correlational analyses. The extended panel's same-family members are not independent, so inference at that level treats families as units; \cref{app:stat-protocol} gives the procedure.
 
\paragraph{Reference tokenizers.} The two off-the-shelf reference tokenizers are held out of both panels. We use them as prediction targets (\cref{sec:results}).
 
\paragraph{Defect exception: three tokenizers without a full byte alphabet.}  Three of the tokenizers we trained do not have all 256 byte values in their vocabulary: these were the ones trained on English-only data with the punct, GPT-4o and Claude pretokenizers. Each has only 224 of the 256 bytes in its vocabulary, and none has a vocabulary entry containing \texttt{\{}. The cause is that the BPE trainer seeds its single-byte vocabulary only from byte values that occur in the tokenizer's own training sample, and merges can only build on entries that exist, so a byte absent from that sample is unrepresentable. 
This is arguably a violation of the defect-free criterion that we stated earlier as necessary for inclusion, but tokenizers often do not contain the full byte- or character-alphabet (for example, byte-fallback is off by default in the HuggingFace Tokenizers library). We thus did not label these tokenizers as defective. 
We note that natural-language and math+code models were retrained with the full byte alphabet for the GPT-4o pretokenization version. The prior defect therefore does not affect the results from those models. The Claude pretokenizer math+code variant  is kept in the math+code panel with the defect flagged, as are the natural-language and math+code models for the punct variant. 
The defect also shifts the affected models' Code BPB. Encoding silently deletes the 32 unrepresentable bytes, so the affected models are never scored on them. One deleted byte, the carriage return, is present in the held-out code data but absent from the math+code training data: it occurs in 3.22\% of sampled held-out StarCoderData documents, and in none of 10{,}001 sampled documents of the math+code training mixture's code half.
A math+code model with a full byte alphabet is therefore scored on carriage returns its training data never contained, while a model whose tokenizer deletes them is not. Consistent with this, the two defect-marked rows of \cref{tab:mc20b-results} print that table's two lowest Code BPB values ($0.365$ and $0.372$), and the third defective build, BPE GPT-4o (eng.), prints $0.3709$ in \cref{tab:cross-regime-code-bpb}, between the other two. These Code BPB values are lower than a full-alphabet counterpart would measure and are not comparable with the other rows; the same English-only GPT-4o configuration retrained with the full byte alphabet has Code BPB $0.399$, which indicates the size of the shift.
The sensitivity of the published correlations to including the punct tokenizer without the full byte alphabet, including the one result whose corrected significance depends on it, is quantified in \cref{app:robustness}.

\begin{table}[t]
\centering
\small
\begin{tabularx}{\textwidth}{@{}l c X X@{}}
\toprule
Panel & $n$ & Membership & Used for \\
\midrule
\primarypanel{} (primary) & \nPrimary{} & One representative per tokenizer family & Aggregate correlations (\cref{tab:aggregate-correlations,tab:aggregate-correlations-fineweb}); per-language regressions (\cref{tab:mixed-effects}) \\
\extendedpanel{} (extended) & \nExtended{} & Every member of every family & Family-mean correlations (\cref{app:robustness}); predictor fits (\cref{app:heldout-reference-prediction}) \\
\mathcodepanel{} (math+code) & \nMCPrimary{} & One representative per family among the custom tokenizers with a math+code model: \nMCInStdPanel{} are also primary-panel members  (\cref{app:tok-inclusion}) & Math+code benchmarks and correlations (\cref{app:mc20b-results}); the math+code columns of \cref{tab:aggregate-correlations} use the \nMCInStdPanel{} primary-panel members only \\
\referencepanel{} (reference) & 2 & The two off-the-shelf tokenizers & Held-out prediction targets (\cref{app:heldout-reference-prediction}); excluded from every ranking and aggregate statistic \\
\bottomrule
\end{tabularx}
\caption{The four tokenizer panels. A tokenizer family groups configurations that share all three design-axis values and differ only in a training hyperparameter or preset; the Fam.\ column of \cref{tab:tokenizer-grid} lists each row's family.}
\label{tab:panel-overview}
\end{table}

\begin{longtable}{@{}p{2.6cm}lccccc p{2.2cm}@{}}
\caption{Configurations for every tokenizer named in this paper: the \nStdAll{} we train plus the 2 off-the-shelf references, the same \nStdAllRefs{} rows as \cref{tab:main-results}.
\textbf{Pretok}: P=Punctuation, G=GPT-4o, G2=GPT-2, C=Claude, R=Right-aligned, Cl=Clean-multi, S=SCRIPT-encoding, W=whitespace only.
\textbf{Data}: B=Balanced, E=English, Co=Code, Hr=Highres (6 langs), Hm=Highmid (21 langs), Eq=Equal-weighted multilingual.
\textbf{Panels}: definitions given in \cref{tab:panel-overview}. \primarypanel{} implies membership in the extended panel too, so P is printed alone; Mx marks a member of the extended math+code roster that is not in the math+code correlation panel (the $^{e}$ rows of \cref{tab:mc20b-results}); \textrm{--}=not a member of any panel.
\textbf{Fam.}: the family index of \cref{app:tok-inclusion}; rows sharing an index are members of one family, and the \nFamilies{} indices partition the \nExtended{}-member extended panel. Rows outside that panel have no family.
\textbf{Note}: an axis not captured by the other columns, e.g.\ the SuperBPE stage-1 base, or a property that disqualifies the row from panel membership.
}
\small
\setlength{\tabcolsep}{4pt}
\label{tab:tokenizer-grid}\\
\textbf{Tokenizer} & \textbf{Algorithm} & \textbf{Pretok} & \textbf{Data} & \textbf{NFC} & \textbf{Panels} & \textbf{Fam.} & \textbf{Note} \\
\midrule
\endfirsthead
\toprule
\textbf{Tokenizer} & \textbf{Algorithm} & \textbf{Pretok} & \textbf{Data} & \textbf{NFC} & \textbf{Panels} & \textbf{Fam.} & \textbf{Note} \\
\midrule
\multicolumn{8}{l}{\small\itshape (continued from previous page)} \\
\midrule
\endhead
\midrule
\multicolumn{8}{r}{\small\itshape continued on next page} \\
\endfoot
\bottomrule
\endlastfoot
Mistral-Nemo & BPE & G & -- & -- & R & -- &  \\
LLaMA-3 & BPE & -- & -- & -- & R & -- &  \\
\midrule
BPE Punct (bal.) & BPE & P & B & \xmark & P, M & 1 &  \\
BPE Punct (eng.) & BPE & P & E & \xmark & P, M & 2 & incomplete byte alphabet \\
BPE GPT-4o (bal.) & BPE & G & B & \xmark & P, M & 3 &  \\
BPE GPT-4o NFC & BPE & G & B & \cmark & P, M & 4 &  \\
BPE GPT-4o (eng.) & BPE & G & E & \xmark & -- & -- & incomplete byte alphabet \\
BPE GPT-4o (eng., full-byte) & BPE & G & E & \xmark & P, M & 5 & full byte alphabet \\
BPE GPT-4o (code) & BPE & G & Co & \xmark & P, M & 6 &  \\
BPE Claude (bal.) & BPE & C & B & \xmark & P, M & 7 &  \\
BPE Claude NFC & BPE & C & B & \cmark & P, M & 8 &  \\
BPE Claude (eng.) & BPE & C & E & \xmark & M & -- & incomplete byte alphabet \\
BPE RightAlign (bal.) & BPE & R & B & \xmark & P, M & 9 &  \\
BPE RightAlign NFC & BPE & R & B & \cmark & P, M & 10 &  \\
Unigram GPT-4o & Unigram & G & B & \xmark & P, M & 11 &  \\
Unigram Claude & Unigram & C & B & \xmark & P, M & 12 &  \\
Unigram RightAlign & Unigram & R & B & \xmark & P, M & 13 &  \\
SuperBPE GPT-4o (bal.) & SuperBPE & G & B & \xmark & P, M & 14 &  \\
PA-BPE NFC GPT-4o & PA-BPE & G & B & \cmark & P, M & 15 &  \\
Unigram GPT-4o (highmid) & Unigram & G & Hm & \xmark & P & 16 & 21 langs \\
Unigram GPT-4o (highres) & Unigram & G & Hr & \xmark & P & 17 & 6 langs \\
MinGram SCRIPT-enc & MinGram & S & B & \cmark & P, M & 18 & SCRIPT enc. \\
MinGram SCRIPT-enc (nl-split) & MinGram & S & B & \cmark & E, \mathcodepanelextended & 18 & SCRIPT enc., splits line breaks \\
BPE SCRIPT-enc GPT-4o & BPE & S & B & \cmark & P, M & 19 & SCRIPT enc., GPT-4o regex \\
BPE GPT-2 (bal.) & BPE & G2 & B & \xmark & P & 20 &  \\
BPE GPT-4o (all-eq.) & BPE & G & Eq & \xmark & P & 21 &  \\
BPE Claude (all-eq.) & BPE & C & Eq & \xmark & P & 22 &  \\
BPE Punct (all-eq.) & BPE & P & Eq & \xmark & P & 23 &  \\
BPE GPT-4o NFC (all-eq.) & BPE & G & Eq & \cmark & P & 24 &  \\
BPE GPT-4o (all-eq., no-rep.) & BPE & G & Eq & \xmark & E & 21 & no repeat-sampling \\
BPE GPT-4o (bal., mf50) & BPE & G & B & \xmark & -- & -- & raised file cap, null control \\
BPE NFC clean (bal.) & BPE & Cl & B & \cmark & P, M & 25 &  \\
BPE NFC clean plus2 (bal.) & BPE & Cl & B & \cmark & E, \mathcodepanelextended & 25 & plus2 apostrophe \\
BPE NFC clean plus3 (bal.) & BPE & Cl & B & \cmark & E, \mathcodepanelextended & 25 & plus3 apostrophe \\
SuperBPE PA-BPE GPT-4o & SuperBPE & G & B & \cmark & E & 26 & PA-BPE base \\
SuperBPE PA-BPE GPT-4o t64k & SuperBPE & G & B & \cmark & E & 26 & PA-BPE base \\
SuperBPE clean C2 (bal.) & SuperBPE & Cl & B & \cmark & P, M & 27 & PA-BPE base, C2 \\
SuperBPE clean C3 (bal.) & SuperBPE & Cl & B & \cmark & E, \mathcodepanelextended & 27 & PA-BPE base, C3 \\
SuperBPE Punct t64k & SuperBPE & P & B & \xmark & E & 28 &  \\
PA-BPE NFC GPT-4o HW & PA-BPE & G & B & \cmark & E & 15 & hybrid window \\
PA-BPE NFC clean & PA-BPE & Cl & B & \cmark & P & 29 &  \\
Unigram GPT-4o tuned & Unigram & G & B & \xmark & E & 11 & tuned hyperparams \\
BPE GPT-4o (highres) & BPE & G & Hr & \xmark & P & 30 & 6 langs \\
BPE GPT-4o (highmid) & BPE & G & Hm & \xmark & P & 31 & 21 langs \\
BPE Whitespace (bal.) & BPE & W & B & \xmark & -- & -- & control \\
Unigram Whitespace tuned (bal.) & Unigram & W & B & \xmark & -- & -- & control, SentencePiece, tuned 
 \end{longtable}

\section{Model Architecture and Training Details}
\label{app:architecture}

\subsection{Architecture}
The model uses the nanochat d24 configuration~\citep{karpathy2025nanochat}: 24 layers, hidden dimension 1536, 12 attention heads (head dimension 128), multi-head attention, context length 2048.
Position information uses RoPE~\citep{su2024roformer} with base frequency $\theta = 100{,}000$.
Activations are squared ReLU ($\text{ReLU}^2$).
Normalization is parameterless RMSNorm.
QK-norm applies a scale of 1.2 to both queries and keys.
Logits are soft-capped: $15 \cdot \tanh(\text{logits}/15)$.
Attention uses Flash Attention~\citep{dao2023flashattention2} on Hopper GPUs with PyTorch SDPA as fallback.

The sliding window pattern ``SSSL'' tiles across layers: three layers attend to $\lceil L/4 \rceil$ tokens (rounded to the Flash Attention tile size), followed by one full-context layer.
Alternating layers include value embeddings: per-token learned vectors of dimension 128, projected to the full key-value dimension (1536) via a learned linear layer, added to the value tensor through an input-dependent gate.
Per-layer learnable scalars control residual stream scaling and skip-connection blending.

\paragraph{Parameter counts.}
All models use a $\sim$128K vocabulary (Mistral-Nemo differs slightly in vocab size at 131K, although $\sim$1K of these are special tokens). Under 128K vocabulary, the parameter counts for our architectures are: 682M for transformer matrices, 197M for input embeddings, 197M output projection, 197M for value embeddings (12 layers $\times$ 128K $\times$ 128), 74 scalar parameters.
This gives us a total of 1.27B parameters. 
Embedding parameters scale with vocabulary; of the two off-the-shelf tokenizers, LLaMA-3 matches the standard 128,256-token vocabulary exactly (checked directly against the trained model weights: 0 parameter difference), while Mistral-Nemo's larger 131,072-token vocabulary adds 12{,}976{,}128 parameters, 1.0\% of the 1.27B total.

\subsection{Training Hyperparameters}
We follow the training recommendations of the nanochat library. We provide a summary of training details below, but refer the reader to the repository for further details and hyperparameter choice justifications. Training is done on 4 NVIDIA GH200 GPUs in \texttt{bfloat16}.
\paragraph{Training budget.}
For the natural-language-focused models, we train on $10.5 \times N_{\text{scale}}$ tokens where $N_{\text{scale}} = $ transformer matrices $+$ output projection $= 879$M. 
This ratio is taken directly from nanochat's scaling law analysis with Kaplan-style parameter counting ($N \propto C^{0.54}$, $D \propto C^{0.49}$).\footnote{See \url{https://github.com/karpathy/nanochat/blob/master/dev/LOG.md}}
This give\crnew{s} a target of 9.23B tokens (8{,}800 steps at batch size $2^{20} = 1{,}048{,}576$).
This budget was held fixed rather than extended, based on a check on five tokenizers (four custom and LLaMA-3) whose models we continued training for a further 10B tokens on the same mixture, roughly twice the compute-optimal budget: val BPB improved for all five with the cross-tokenizer ranking preserved, but the math and code generation benchmarks stayed near floor (MBPP pass@1 at most $0.070$ across the five continued models) with unstable cross-tokenizer rankings. This motivated the separate, dedicated training of the math+code models. 

\paragraph{Optimizer.}
Weight matrices use the Muon optimizer~\citep{jordan2025muon}; embeddings, projections and scalars use AdamW.
Learning rates are transferred across widths via $\mu$P~\citep{yang2022tensor}. 
Base learning rates: 0.02 (Muon), 0.30 (input embeddings), 0.008 (output projection), 0.50 (scalar parameters).
AdamW LRs scaled by $(d_{\text{model}} / 768)^{-1} = 0.5$ ($\mu$P~\citep{yang2022tensor}).
All LRs scaled by $\sqrt{B / B_{\text{ref}}} = \sqrt{2} \approx 1.414$ where $B_{\text{ref}} = 524{,}288$. We use \texttt{torch.compile(dynamic=False)} and gradient accumulation of 8 micro-batches. \cref{tab:optimizer} lists the resulting effective learning rates and weight decay per parameter group.

\begin{table}[h]
\centering
\caption{Effective learning rates and optimizer settings per parameter group.}
\label{tab:optimizer}
\small
\begin{tabular}{lllcl}
\toprule
\textbf{Group} & \textbf{Optimizer} & \textbf{Eff.\ LR} & \textbf{Betas} & \textbf{WD} \\
\midrule
Transformer matrices & Muon & 0.0283 & mom.\ schedule & 0.050$\to$0 \\
Input embeddings & AdamW & 0.2121 & (0.8, 0.995) & 0.001 \\
Output projection & AdamW & 0.0057 & (0.8, 0.96) & 0.01 \\
Value embeddings & AdamW & 0.1061 & (0.8, 0.995) & 0.01 \\
Residual scalars & AdamW & 0.0071 & (0.8, 0.95) & 0.05 \\
Skip scalars & AdamW & 0.707 & (0.96, 0.95) & 0.0 \\
\bottomrule
\end{tabular}
 \end{table}

\paragraph{Schedule.}
We use a linear LR warm-up over 40 steps, holding it constant until 35\% of training. LR then decays linearly over the remaining 65\% to 5\% of peak.
Muon momentum: 0.85$\to$0.97 over 400 steps, constant at 0.97, then 0.97$\to$0.90 during warmdown.
Weight decay: $\lambda_{\text{ref}} \cdot \sqrt{B/B_{\text{ref}}} \cdot (D_{\text{ref}}/D) = 0.050$, with cosine decay to zero ($T_{\text{epoch}}$ framework, which scales weight decay to hold $T_{\text{epoch}} = B/(\eta \lambda D)$ constant).

\subsection{Training Data Composition}
\label{app:data}

\subsubsection{Natural Language-Focused Track}
Our model training data consists of 36.9\% English web (FineWeb-Edu), 33.4\% multilingual in 30 languages spanning 11 scripts (FineWeb2), 16\% math (FineMath), and 13.6\% code (StarCoderData), where percentages are computed using text bytes.
Note that the multilingual portion creates a 266$\times$ data scarcity gradient from Russian (18\% of bytes) to Tamil (0.1\%), mimicking the drastic differences between languages' representations observed in most real-world datasets. Documents are drawn one at a time from a source chosen with probability proportional to its weight; the per-language weights are proportional to estimated character counts in the source FineWeb2 corpus.  This produces a data scarcity gradient from Russian (18.1\% of the mixture's bytes, 531K docs) to Tamil (0.1\%, 1.8K docs).
A language's byte share scales with its weight times its mean document length, so the mixture's byte shares deviate from the source corpus's: Russian is 54.2\% of the multilingual bytes vs. a 32.9\% share of these 30 languages' bytes in the FineWeb2, and Chinese 3.9\% vs. 11.8\%. 
\cref{tab:data-composition} gives the overall domain composition (English, multilingual, math, code).
The full per-language breakdown is in \cref{tab:lang-breakdown}.

\begin{table}[h]
\centering
\caption{LM training data composition (5M documents, 25\,GB UTF-8 text).}
\label{tab:data-composition}
\small
\begin{tabular}{llrr}
\toprule
\textbf{Domain} & \textbf{Source} & \textbf{Bytes} & \textbf{\%} \\
\midrule
English web           & FineWeb-Edu    & 9.2\,GB & 36.9 \\
Multilingual (30 langs)& FineWeb2       & 8.3\,GB & 33.4 \\
Math                  & FineMath-4plus & 4.0\,GB & 16.1 \\
Code (Python)         & StarCoderData  & 2.5\,GB & 10.0 \\
Code (JavaScript)     & StarCoderData  & 0.9\,GB &  3.6 \\
\midrule
\textbf{Total}        &                & \textbf{25.0\,GB} & 100.0 \\
\bottomrule
\end{tabular}
 \end{table}

\begin{table}[h]
\centering
\caption{Multilingual training data by language, sorted by byte share.
11 writing systems, 11+ language families. Data scarcity spans 266$\times$
from Russian to Tamil.}
\label{tab:lang-breakdown}
\scriptsize
\begin{tabular}{llcrrc}
\toprule
\textbf{Language} & \textbf{Code} & \textbf{Script} & \textbf{GB} & \textbf{\%} & \textbf{Docs} \\
\midrule
Russian & \texttt{rus\_Cyrl} & Cyrillic & 4.521 & 18.1 & 531,387 \\
Spanish & \texttt{spa\_Latn} & Latin & 0.423 & 1.7 & 122,231 \\
German & \texttt{deu\_Latn} & Latin & 0.353 & 1.4 & 118,340 \\
French & \texttt{fra\_Latn} & Latin & 0.364 & 1.5 & 103,496 \\
Chinese & \texttt{cmn\_Hani} & CJK & 0.326 & 1.3 & 95,805 \\
Japanese & \texttt{jpn\_Jpan} & CJK & 0.204 & 0.8 & 63,316 \\
Italian & \texttt{ita\_Latn} & Latin & 0.165 & 0.7 & 59,676 \\
Portuguese & \texttt{por\_Latn} & Latin & 0.189 & 0.8 & 59,460 \\
Turkish & \texttt{tur\_Latn} & Latin & 0.152 & 0.6 & 47,074 \\
Indonesian & \texttt{ind\_Latn} & Latin & 0.236 & 0.9 & 45,970 \\
Polish & \texttt{pol\_Latn} & Latin & 0.107 & 0.4 & 43,902 \\
Ukrainian & \texttt{ukr\_Cyrl} & Cyrillic & 0.211 & 0.8 & 34,088 \\
Dutch & \texttt{nld\_Latn} & Latin & 0.071 & 0.3 & 32,404 \\
Romanian & \texttt{ron\_Latn} & Latin & 0.097 & 0.4 & 29,203 \\
Arabic & \texttt{arb\_Arab} & Arabic & 0.168 & 0.7 & 27,292 \\
Hungarian & \texttt{hun\_Latn} & Latin & 0.099 & 0.4 & 25,066 \\
Vietnamese & \texttt{vie\_Latn} & Latin & 0.109 & 0.4 & 21,306 \\
Czech & \texttt{ces\_Latn} & Latin & 0.077 & 0.3 & 21,696 \\
Greek & \texttt{ell\_Grek} & Greek & 0.108 & 0.4 & 17,672 \\
Thai & \texttt{tha\_Thai} & Thai & 0.087 & 0.3 & 11,639 \\
Finnish & \texttt{fin\_Latn} & Latin & 0.034 & 0.1 & 11,721 \\
Slovak & \texttt{slk\_Latn} & Latin & 0.028 & 0.1 & 10,015 \\
Bulgarian & \texttt{bul\_Cyrl} & Cyrillic & 0.052 & 0.2 & 9,513 \\
Korean & \texttt{kor\_Hang} & Hangul & 0.029 & 0.1 & 8,263 \\
Croatian & \texttt{hrv\_Latn} & Latin & 0.026 & 0.1 & 8,395 \\
Catalan & \texttt{cat\_Latn} & Latin & 0.014 & 0.1 & 5,087 \\
Hindi & \texttt{hin\_Deva} & Devanagari & 0.032 & 0.1 & 4,928 \\
Hebrew & \texttt{heb\_Hebr} & Hebrew & 0.025 & 0.1 & 4,665 \\
Bengali & \texttt{ben\_Beng} & Bengali & 0.024 & 0.1 & 3,525 \\
Tamil & \texttt{tam\_Taml} & Tamil & 0.017 & 0.1 & 1,831 \\
\bottomrule
\end{tabular}
 \end{table}

\subsubsection{Math+Code-Focused Track}
We build the math+code mixture from two sources: MegaMath-Web-Pro (LLM360 MegaMath; $\sim$50\,GB text, $\sim$14M documents) for math, and The Stack v2's educational subset for code, restricted to seven languages weighted proportionally to their share of the source data: JavaScript (12.7\,GB, 3.4M docs), Java (11.0\,GB, 3.4M docs), Python (11.0\,GB, 2.3M docs), C++ (7.1\,GB, 1.4M docs), TypeScript (3.5\,GB, 1.3M docs), Go (3.4\,GB, 0.8M docs), and Rust (1.3\,GB, 0.2M docs), together $\sim$50\,GB and $\sim$13M documents. 
Eight further \texttt{stackv2-edu} languages (C, C\#, PHP, Ruby, Shell, SQL, Swift, Markdown) are excluded as less relevant to the code benchmarks of \cref{sec:evaluation}. The combined mixture is $\sim$100\,GB of text over $\sim$27M documents, 50\% math and 50\% code by bytes; documents are sampled with the same weighted-random mechanism as the natural-language mixture above, so any prefix of the mixture keeps this ratio. We use a $\sim$20B-token prefix to train the math+code models.

\section{Evaluation Details}\label{app:eval-harness}

\subsection{Downstream Evaluation Configurations}
\paragraph{Loglikelihood scoring.} BLiMP and MultiBLiMP are minimal-pair benchmarks: each item pairs an acceptable sentence with a minimally differing unacceptable one, the model assigns each full sentence its total log-likelihood, and the item counts as correct when the acceptable sentence receives the higher total. Accuracy is the fraction of items scored correctly; no generation is involved. Both benchmarks score the sentences from an empty context: the beginning-of-sequence token is prepended and every token of the sentence is scored (the BOS-prefixed convention), matching how documents are packed during training, where every document is itself preceded by the beginning-of-sequence token. The totals are unnormalized log-likelihoods, so sentence pairs whose members tokenize to different lengths are compared on unequal token counts; this is the lm-evaluation-harness reference convention for these tasks, and \cref{app:cross-scale-ranking} notes one consequence for cross-scale BLiMP comparisons.

\paragraph{Generation harness.} GSM8K, HumanEval, and MBPP (\cref{app:mc20b-results}) use the Language Model Evaluation Harness~\citep[v0.4.11;][]{gao2024lmeval}: GSM8K is 8-shot chain-of-thought with exact-match scoring on the full 1{,}319-problem test set; HumanEval is pass@1 under greedy 0-shot decoding; MBPP is pass@1 under greedy 3-shot decoding. Generation heals the prompt boundary: where the prompt's own tokenization does not end on a token boundary the tokenizer would itself produce there, the decoder drops the affected trailing prompt token(s) and constrains generation until their content is re-emitted, so scoring never starts from a token sequence absent from training.

\paragraph{Bits-per-byte.} $\text{BPB} = \sum_i \text{NLL}_i / (\ln 2 \cdot B)$, where $\text{NLL}_i$ is the model's negative log-likelihood, in nats, for scored token $i$; dividing by $\ln 2$ converts nats to bits; and $B$ is the UTF-8 byte length of the original input text. There are several subtle design choices in this metric that must be taken into account. When the model's context window truncates a document, the document contributes its original byte length scaled by the scored fraction of the decoded text (the byte length of the decoded scored tokens divided by the byte length of the full decoded text).  Normalizing by byte length rather than by token count makes the metric comparable across tokenizers regardless of differences in vocabulary or compression rates. A truncated document contributes negative log-likelihood terms for its
scored tokens only. If such a document contributed its full original byte length
to $B$, those terms would be divided by the byte length of the whole document,
and the result would be lower than the bits per byte of the scored span.
Multiplying the original byte length by the scored fraction, that is, by the byte
length of the decoded scored tokens divided by the byte length of the full
decoded text, makes the numerator and the denominator refer to the same span of
text. Using the original text's byte length, rather than the byte length of the
text recovered by decoding, also holds $B$ fixed across tokenizers for a given
document. Encoding followed by decoding does not always return the input text: a
Unicode normalizer applied during encoding, or a character absent from the
tokenizer's alphabet, changes the decoded text and therefore its byte length. If
$B$ were the decoded byte length, two tokenizers with different decoded lengths
for the same document would give BPB values differing by the ratio of those two
lengths, even if the summed negative log-likelihood were identical. That
difference would be a property of the encoding procedure, not of the model's
predictions.

\subsection{Statistical Testing Protocol}\label{app:stat-protocol}
 
The tokenizer panels are designed grids, not random samples, and two sources of statistical dependence follow from that design. First, members of one tokenizer family are close to duplicates, so treating every tokenizer as an independent observation would overstate the amount of evidence and understate uncertainty; the family partition of \cref{app:tok-inclusion} is the unit that addresses this. Second, the per-language analyses observe each tokenizer once per language, so the rows contributed by one tokenizer are correlated with each other; clustering standard errors by tokenizer addresses this. Throughout, a \emph{standard error} is the expected size of an estimate's variation if the experiment were repeated, tests are two-sided, and significance means a Benjamini--Hochberg-adjusted $p<0.05$ (\cref{sec:evaluation}). \Cref{tab:inference-layers} maps each analysis to the panel it runs on and the uncertainty method it uses; the paragraphs below follow the same order.
 
\begin{table}[t]
\centering
\small
\newcolumntype{Y}[1]{>{\raggedright\arraybackslash\hsize=#1\hsize}X}

\begin{tabularx}{\textwidth}{@{}Y{1.3}Y{0.7}Y{0.7}Y{1.3}@{}}
\toprule
Analysis & Panel ($n$) & Point estimate & Uncertainty and correction \\
\midrule
Aggregate correlation (\cref{tab:aggregate-correlations,tab:aggregate-correlations-fineweb}) & \primarypanel{} (\nPrimary); math+code columns \nMCInStdPanel{} & Spearman $\rho$ & Analytic $p$-value; BH correction within each table's printed cells \\
Family-mean correlation (\cref{app:robustness}) & \extendedpanel{} (\nExtended{} tokenizers in \nFamilies{} families) & Spearman $\rho$ on family means & Permutation $p$-value over intact families; BH correction within the artifact grid \\
Per-language regression (\cref{tab:mixed-effects}) & \primarypanel{} $\times$ languages & Mixed-effects $\beta$ & Tokenizer-clustered standard errors; BH correction per outcome \\
Held-out prediction (\cref{app:heldout-reference-prediction}) & Fits on \extendedpanel{} and \mathcodepanel{} & Predicted downstream value & 90\% prediction interval calibrated on cross-validation errors \\
\bottomrule
\end{tabularx}
\caption{The four inference layers, the panel each runs on, and the uncertainty method each uses.}
\label{tab:inference-layers}
\end{table}
 
\paragraph{Primary panel: analytic Spearman tests.} On the primary panel, Spearman correlations are tested with the analytic $p$-value, adjusted within the correction families listed under Multiple comparisons below. \Cref{app:tok-inclusion} states why the panel's construction makes the independence approximation reasonable, and \cref{app:robustness} probes the remaining shared-design-axis structure directly.
 
\paragraph{Extended panel: family-mean permutation tests.} At the extended level, each family is first averaged to one point before ranking, so same-family tokenizers do not count as independent data points. Significance is then assessed with a permutation test: the outcome values are reshuffled among families $B = 10{,}000$ times with a fixed seed, keeping each family intact, and the permutation $p$-value is the fraction of reshuffles whose $|\rho|$ reaches the observed value, computed as $(h+1)/(B+1)$, where $h$ counts the qualifying reshuffles, so the $p$-value is never exactly zero. The statistic is not computed in analyses using less than 5 families.
 
\paragraph{Held-out prediction intervals.} The family partitions given above also define the cross-validation folds for the held-out prediction analysis (\cref{app:heldout-reference-prediction}). Concretely, each fold holds out one whole tokenizer family. This means that no regression model is scored on a tokenizer that belongs to the same family as one of the tokenizers in its training data. 
To compute 90\% prediction intervals, we take the predicted value and add and subtract (to get the intervals upper and lower bounds) a multiple of the root-mean-squared error computed from cross validation. In plain terms, this root-mean-squared error is the typical size of the prediction errors observed in cross-validation. For the multiplier, we use values corresponding to a 90\% from the $t$ distribution rather than the normal distribution, which gives a wider interval, accounting for that error size being itself estimated from a small number of folds. 
 
\paragraph{Per-language regressions: two fits, clustered standard errors.} The cells of \cref{tab:mixed-effects} combine two fits. The printed $\beta$ comes from the mixed-effects regression: outcome $\sim$ metric $+\ (1\,|\,$language$)$, with the metric standardized. The printed standard error and the $p$-values come from a second fit of the same within-language quantity: ordinary least squares with one fixed intercept per language. In this second fit, we cluster standard errors for data points produced by the same tokenizer ($G = 29$ clusters; each family contributes one tokenizer to this panel, so tokenizer-level and family-level clustering coincide here). This choice models the fact that residuals contributed by the same tokenizer across different languages are correlated, \emph{not} independent. The second fit is needed because the mixed-effects standard errors treat those residuals as independent, which results in an unrealistically small estimate of standard error.\footnote{Clustering increases the estimated standard errors by factors of 1.05 to 2.77. Using the clustered estimates changes the significance status of one of the 18 metric-outcome cells: fertility against MultiBLiMP accuracy, adjusted $p = 0.00026$ unclustered vs. $0.063$ clustered.} The two fits' coefficients agree at the printed precision. 
For $p$-values computed using the clustered standard error estimates, we use a normal reference distribution. Using the more conservative $t$ distribution as our reference distribution changes results minimally. Both inference variants are in the released artifacts.
 
\paragraph{Multiple comparisons.} A Benjamini--Hochberg correction is a protocol for correcting p-values when multiple comparisons are being performed, adjusting for the higher likelihood of at-chance encountering a false positive. It is applied within the following correction families:
\begin{itemize}
    \item the 98 printed cells of \cref{tab:aggregate-correlations} (14 metrics $\times$ 7 downstream columns);
    \item the 98 printed cells of \cref{tab:aggregate-correlations-fineweb}, the parallel grid under FineWeb measurement;
    \item the 160-cell grid of the released correlation artifact (20 metrics $\times$ 8 downstream targets); this grid is the correction family behind the natural-language code BPB correlations quoted in \cref{app:mc20b-results}, a downstream column computed but not printed in the paper's tables;
    \item the 9 metrics of each outcome of \cref{tab:mixed-effects}, corrected per outcome;
    \item the 38 code-structure tests of \cref{app:mc20b-results};
    \item the per-example families of \cref{app:external-per-example}, one per text basis and task, of 7 to 9 tests each.
\end{itemize}
One association is tested in two of these families on two rosters: AST boundary alignment against MBPP pass@1 is $\rho = 0.61$ over the \nMCInStdPanel{} primary-panel members with a math+code model (\cref{tab:aggregate-correlations}) and $\rho = 0.65$ over the full \nMCPrimary-member math+code panel (\cref{app:mc20b-results}); the two values describe the same relationship on nested rosters. The line-break statistics of \cref{app:newline-mechanism} and the cross-model rank correlation of \cref{app:cross-regime-code-bpb} are pre-specified single comparisons rather than scans over a metric grid, and are reported without correction: the former as raw $p$-values, the latter as a bootstrap confidence interval, a range of values obtained by recomputing the statistic on many resamples of the data, expected to contain the true value in 95\% of repetitions.
 
\paragraph{Robustness and disclosure.} Panel membership was frozen before any statistic was computed, and results are reported regardless of direction. Each correlation discussed in \cref{sec:results} has a standing robustness artifact: single-member leave-one-out over the primary panel, plus recomputation with each design-axis group removed in turn (\cref{app:robustness}). Where a leverage check removes members, as in the operator-isolation paragraph of \cref{app:mc20b-results}, the full-panel result is reported first and the removal is a labeled diagnostic, not a substitution.

\section{Intrinsic Evaluation Results}\label{app:intrinsic-results}
\cref{tab:intrinsic-flores,tab:intrinsic-fineweb} report the intrinsic metric values for the trained tokenizers and for the two off-the-shelf references, measured on FLORES+ and on FineWeb-Edu/FineWeb2 respectively; the two tables' length-unit conventions differ and are stated in their captions. The largest notable change is in each corpus's reported compression rate: in FLORES+, the denominator is the number of lines, while in FineWeb-Edu/FineWeb2 it is the number of bytes, greatly changing the magnitude of the values across the different measurement corpora. 

\begin{table*}[t]
\centering
\caption{Intrinsic tokenizer evaluation results computed on a subset of the FLORES+ validation set, in particular on the subset of languages in our training data. Code and math metrics are computed on StarCoder and math data respectively. Length-unit-dependent metrics use $u=\texttt{lines}$, with one exception: Fertility keeps its default $u=\texttt{words}$ in both this table and \cref{tab:intrinsic-fineweb} (\cref{tab:metric-summary}). Compression is therefore measured in \emph{lines} per token and is not comparable with the Compression column of \cref{tab:intrinsic-fineweb}, which uses $u=\texttt{bytes}$. Op Isolation is the math-corpus domain, not the value pooled over prose, code and math. Tokenizers are the \nIntrinsicAll{} members of the \cref{tab:main-results}, plus the two off-the-shelf reference tokenizers (marked with $^\dagger$), excluded from every ranking and aggregate statistic except the cross-scale check of \cref{tab:cross-scale-ranking}.}
\label{tab:intrinsic-flores}
\small
\adjustbox{max width=\textwidth}{%
\begin{tabular}{@{}lcccccccccc@{}}
\toprule
\textbf{Tokenizer} & \textbf{Fertility} $\downarrow$ & \textbf{Compression} $\uparrow$ & \textbf{Gini} $\downarrow$ & \textbf{Vocab CoV} $\downarrow$ & \textbf{$\bar H_2$} $\uparrow$ & \textbf{Bigram $\eta$} $\uparrow$ & \textbf{Char Split} $\downarrow$ & \textbf{AST Align} $\uparrow$ & \textbf{Digit F$_1$} $\uparrow$ & \textbf{Op Isolation} $\uparrow$ \\
\midrule
Mistral-Nemo$^\dagger$ & 4.78 & 0.0241 & 0.100 & 0.435 & 0.554 & 0.888 & 0.008 & 0.535 & 0.483 & 0.987 \\
LLaMA-3$^\dagger$ & 5.26 & 0.0185 & 0.238 & 0.447 & 0.505 & 0.822 & 0.137 & 0.511 & 0.643 & 0.987 \\
\midrule
BPE Punct (bal.) & 4.78 & 0.0215 & 0.196 & 0.495 & 0.535 & 0.843 & 0.025 & 1.000 & 0.662 & 1.000 \\
BPE Punct (eng.) & 9.39 & 0.0131 & 0.267 & 0.721 & 0.501 & 0.767 & 0.247 & 1.000 & 0.671 & 1.000 \\
BPE GPT-4o (bal.) & 4.34 & 0.0235 & 0.128 & 0.491 & 0.561 & 0.890 & 0.031 & 0.545 & 0.643 & 0.987 \\
BPE GPT-4o NFC & 4.34 & 0.0236 & 0.126 & 0.491 & 0.561 & 0.891 & 0.031 & 0.545 & 0.643 & 0.987 \\
BPE GPT-4o (eng.) & 9.57 & 0.0130 & 0.269 & 0.735 & 0.499 & 0.772 & 0.287 & 0.658 & 0.643 & 0.990 \\
BPE GPT-4o (eng., full-byte) & 9.57 & 0.0130 & 0.269 & 0.735 & 0.499 & 0.772 & 0.287 & 0.683 & 0.643 & 0.990 \\
BPE GPT-4o (code) & 7.59 & 0.0139 & 0.256 & 0.676 & 0.513 & 0.780 & 0.243 & 0.516 & 0.643 & 0.987 \\
BPE Claude (bal.) & 4.33 & 0.0237 & 0.125 & 0.486 & 0.560 & 0.889 & 0.030 & 0.748 & 0.643 & 0.997 \\
BPE Claude NFC & 4.32 & 0.0237 & 0.123 & 0.486 & 0.560 & 0.890 & 0.030 & 0.748 & 0.643 & 0.997 \\
BPE Claude (eng.) & 9.41 & 0.0132 & 0.264 & 0.734 & 0.504 & 0.774 & 0.274 & 0.787 & 0.643 & 0.997 \\
BPE RightAlign (bal.) & 4.34 & 0.0235 & 0.128 & 0.487 & 0.561 & 0.889 & 0.031 & 0.545 & 1.000 & 0.987 \\
BPE RightAlign NFC & 4.34 & 0.0236 & 0.126 & 0.487 & 0.561 & 0.890 & 0.031 & 0.545 & 1.000 & 0.987 \\
Unigram GPT-4o & 4.27 & 0.0192 & 0.121 & 0.393 & 0.324 & 0.770 & 0.049 & 0.724 & 0.483 & 1.000 \\
Unigram Claude & 4.30 & 0.0192 & 0.120 & 0.380 & 0.333 & 0.773 & 0.054 & 0.794 & 0.483 & 1.000 \\
Unigram RightAlign & 4.27 & 0.0192 & 0.121 & 0.394 & 0.324 & 0.770 & 0.049 & 0.724 & 0.483 & 1.000 \\
SuperBPE GPT-4o (bal.) & 4.40 & 0.0234 & 0.158 & 0.541 & 0.629 & 0.899 & 0.037 & 0.416 & 0.643 & 0.808 \\
PA-BPE NFC GPT-4o & 3.99 & 0.0275 & 0.017 & 0.259 & 0.540 & 0.911 & 0.015 & 0.605 & 0.633 & 0.990 \\
Unigram GPT-4o (highmid) & 4.81 & 0.0161 & 0.224 & 0.492 & 0.340 & 0.740 & 0.186 & 0.724 & 0.483 & 1.000 \\
Unigram GPT-4o (highres) & 6.48 & 0.0141 & 0.201 & 0.618 & 0.335 & 0.724 & 0.221 & 0.716 & 0.483 & 1.000 \\
MinGram SCRIPT-enc & 4.22 & 0.0244 & 0.119 & 0.478 & 0.542 & 0.891 & 0.001 & 0.748 & 0.637 & 0.997 \\
MinGram SCRIPT-enc (nl-split) & 4.22 & 0.0244 & 0.119 & 0.478 & 0.542 & 0.891 & 0.001 & 0.748 & 0.637 & 0.997 \\
BPE SCRIPT-enc GPT-4o & 4.31 & 0.0237 & 0.123 & 0.490 & 0.557 & 0.890 & 0.001 & 0.546 & 0.643 & 0.987 \\
BPE GPT-2 (bal.) & 4.76 & 0.0216 & 0.195 & 0.493 & 0.546 & 0.848 & 0.026 & 0.749 & 0.662 & 0.997 \\
BPE GPT-4o (all-eq.) & 4.07 & 0.0264 & 0.050 & 0.267 & 0.547 & 0.910 & 0.015 & 0.546 & 0.643 & 0.987 \\
BPE Claude (all-eq.) & 4.06 & 0.0265 & 0.050 & 0.259 & 0.545 & 0.909 & 0.015 & 0.748 & 0.643 & 0.997 \\
BPE Punct (all-eq.) & 4.60 & 0.0230 & 0.166 & 0.343 & 0.528 & 0.842 & 0.014 & 1.000 & 0.660 & 1.000 \\
BPE GPT-4o NFC (all-eq.) & 4.07 & 0.0265 & 0.048 & 0.267 & 0.547 & 0.912 & 0.015 & 0.546 & 0.643 & 0.987 \\
BPE GPT-4o (all-eq., no-rep.) & 4.07 & 0.0264 & 0.050 & 0.267 & 0.547 & 0.910 & 0.015 & 0.546 & 0.643 & 0.987 \\
BPE GPT-4o (bal., mf50) & 4.34 & 0.0235 & 0.128 & 0.491 & 0.561 & 0.890 & 0.031 & 0.545 & 0.643 & 0.987 \\
BPE NFC clean (bal.) & 4.33 & 0.0234 & 0.137 & 0.507 & 0.557 & 0.881 & 0.032 & 0.749 & 0.483 & 0.997 \\
BPE NFC clean plus2 (bal.) & 4.35 & 0.0235 & 0.122 & 0.498 & 0.557 & 0.883 & 0.030 & 0.749 & 0.483 & 0.997 \\
BPE NFC clean plus3 (bal.) & 4.35 & 0.0235 & 0.122 & 0.499 & 0.557 & 0.883 & 0.030 & 0.751 & 0.483 & 0.997 \\
SuperBPE PA-BPE GPT-4o & 3.99 & 0.0277 & 0.036 & 0.313 & 0.598 & 0.926 & 0.016 & 0.446 & 0.643 & 0.845 \\
SuperBPE PA-BPE GPT-4o t64k & 4.02 & 0.0271 & 0.051 & 0.355 & 0.610 & 0.926 & 0.018 & 0.425 & 0.643 & 0.815 \\
SuperBPE clean C2 (bal.) & 4.09 & 0.0269 & 0.036 & 0.328 & 0.545 & 0.908 & 0.016 & 0.734 & 0.483 & 0.997 \\
SuperBPE clean C3 (bal.) & 3.97 & 0.0277 & 0.040 & 0.326 & 0.615 & 0.928 & 0.016 & 0.517 & 0.643 & 0.838 \\
SuperBPE Punct t64k & 4.47 & 0.0232 & 0.169 & 0.535 & 0.658 & 0.898 & 0.031 & 0.445 & 0.650 & 0.391 \\
PA-BPE NFC GPT-4o HW & 3.95 & 0.0274 & 0.030 & 0.252 & 0.542 & 0.913 & 0.014 & 0.566 & 0.643 & 0.990 \\
PA-BPE NFC clean & 4.12 & 0.0266 & 0.019 & 0.276 & 0.540 & 0.898 & 0.015 & 0.752 & 0.483 & 0.997 \\
Unigram GPT-4o tuned & 4.19 & 0.0192 & 0.125 & 0.402 & 0.327 & 0.776 & 0.051 & 0.721 & 0.483 & 0.997 \\
BPE GPT-4o (highres) & 7.52 & 0.0143 & 0.309 & 0.656 & 0.484 & 0.772 & 0.268 & 0.543 & 0.643 & 0.987 \\
BPE GPT-4o (highmid) & 5.07 & 0.0180 & 0.302 & 0.526 & 0.470 & 0.807 & 0.205 & 0.545 & 0.643 & 0.987 \\
BPE Whitespace (bal.) & 4.39 & 0.0233 & 0.138 & 0.501 & 0.683 & 0.899 & 0.036 & 0.409 & 0.630 & 0.828 \\
\bottomrule
\end{tabular}}
 \end{table*}

\begin{table*}[t]
\centering
\caption{Intrinsic tokenizer evaluation results computed on a subset of FineWeb-Edu and FineWeb2. Concretely, we sample 1k documents per language for each of the languages present in our training data. Length-unit-dependent metrics use $u=\texttt{bytes}$, with one exception: Fertility keeps its default $u=\texttt{words}$ in both this table and \cref{tab:intrinsic-flores} (\cref{tab:metric-summary}). The value after Fertility's $\pm$ is the standard deviation of the per-document tokens-per-word ratio across the sampled documents, which \cref{tab:intrinsic-flores} does not print. Compression is measured in \emph{bytes} per token and is not comparable with the Compression column of \cref{tab:intrinsic-flores}, which uses $u=\texttt{lines}$. Tokenizers are the same \nIntrinsicAll{} members as \cref{tab:intrinsic-flores}, plus the two off-the-shelf reference tokenizers (marked with $^\dagger$), excluded from every ranking and aggregate statistic except the cross-scale check of \cref{tab:cross-scale-ranking}}.
\label{tab:intrinsic-fineweb}
\small
\adjustbox{max width=\textwidth}{%
\begin{tabular}{@{}lccccccc@{}}
\toprule
\textbf{Tokenizer} & \textbf{Fertility} $\downarrow$ & \textbf{Compression} $\uparrow$ & \textbf{Gini} $\downarrow$ & \textbf{Vocab CoV} $\downarrow$ & \textbf{$\bar H_2$} $\uparrow$ & \textbf{Bigram $\eta$} $\uparrow$ & \textbf{Char Split} $\downarrow$ \\
\midrule
Mistral-Nemo$^\dagger$ & 4.82 {\scriptsize $\pm$13.94} & 4.2326 & 0.128 & 0.396 & 0.572 & 0.787 & 0.010 \\
LLaMA-3$^\dagger$ & 5.15 {\scriptsize $\pm$12.40} & 3.5226 & 0.161 & 0.388 & 0.542 & 0.734 & 0.105 \\
\midrule
BPE Punct (bal.) & 4.73 {\scriptsize $\pm$11.64} & 3.9624 & 0.111 & 0.417 & 0.533 & 0.760 & 0.028 \\
BPE Punct (eng.) & 8.90 {\scriptsize $\pm$26.00} & 2.4434 & 0.135 & 0.534 & 0.519 & 0.669 & 0.226 \\
BPE GPT-4o (bal.) & 4.36 {\scriptsize $\pm$11.29} & 4.2847 & 0.132 & 0.407 & 0.578 & 0.801 & 0.033 \\
BPE GPT-4o NFC & 4.36 {\scriptsize $\pm$11.29} & 4.2847 & 0.132 & 0.407 & 0.578 & 0.801 & 0.033 \\
BPE GPT-4o (eng.) & 8.95 {\scriptsize $\pm$26.20} & 2.4458 & 0.140 & 0.537 & 0.523 & 0.678 & 0.259 \\
BPE GPT-4o (eng., full-byte) & 8.95 {\scriptsize $\pm$26.21} & 2.4455 & 0.140 & 0.537 & 0.523 & 0.678 & 0.259 \\
BPE GPT-4o (code) & 6.93 {\scriptsize $\pm$16.48} & 2.6289 & 0.103 & 0.501 & 0.538 & 0.681 & 0.200 \\
BPE Claude (bal.) & 4.42 {\scriptsize $\pm$11.55} & 4.2538 & 0.132 & 0.414 & 0.544 & 0.791 & 0.032 \\
BPE Claude NFC & 4.42 {\scriptsize $\pm$11.55} & 4.2538 & 0.132 & 0.414 & 0.544 & 0.791 & 0.032 \\
BPE Claude (eng.) & 8.89 {\scriptsize $\pm$26.05} & 2.4553 & 0.135 & 0.540 & 0.522 & 0.677 & 0.246 \\
BPE RightAlign (bal.) & 4.36 {\scriptsize $\pm$11.29} & 4.2847 & 0.132 & 0.405 & 0.578 & 0.801 & 0.033 \\
BPE RightAlign NFC & 4.36 {\scriptsize $\pm$11.29} & 4.2847 & 0.132 & 0.405 & 0.578 & 0.801 & 0.033 \\
Unigram GPT-4o & 4.40 {\scriptsize $\pm$9.11} & 3.5065 & 0.175 & 0.365 & 0.326 & 0.684 & 0.044 \\
Unigram Claude & 4.51 {\scriptsize $\pm$9.64} & 3.4771 & 0.173 & 0.365 & 0.335 & 0.679 & 0.049 \\
Unigram RightAlign & 4.40 {\scriptsize $\pm$9.11} & 3.5064 & 0.175 & 0.365 & 0.326 & 0.684 & 0.044 \\
SuperBPE GPT-4o (bal.) & 4.41 {\scriptsize $\pm$11.46} & 4.2975 & 0.128 & 0.451 & 0.665 & 0.818 & 0.038 \\
PA-BPE NFC GPT-4o & 4.20 {\scriptsize $\pm$12.48} & 4.8100 & 0.146 & 0.259 & 0.562 & 0.826 & 0.017 \\
Unigram GPT-4o (highmid) & 4.93 {\scriptsize $\pm$9.03} & 3.0457 & 0.179 & 0.400 & 0.338 & 0.660 & 0.154 \\
Unigram GPT-4o (highres) & 5.89 {\scriptsize $\pm$11.19} & 2.6422 & 0.163 & 0.516 & 0.334 & 0.638 & 0.200 \\
MinGram SCRIPT-enc & 4.33 {\scriptsize $\pm$11.42} & 4.3649 & 0.130 & 0.403 & 0.526 & 0.792 & 0.000 \\
MinGram SCRIPT-enc (nl-split) & 4.33 {\scriptsize $\pm$11.42} & 4.3654 & 0.130 & 0.403 & 0.526 & 0.792 & 0.000 \\
BPE SCRIPT-enc GPT-4o & 4.35 {\scriptsize $\pm$11.28} & 4.3009 & 0.132 & 0.402 & 0.572 & 0.801 & 0.000 \\
BPE GPT-2 (bal.) & 4.70 {\scriptsize $\pm$11.62} & 4.0035 & 0.112 & 0.411 & 0.546 & 0.765 & 0.029 \\
BPE GPT-4o (all-eq.) & 4.18 {\scriptsize $\pm$12.10} & 4.7119 & 0.140 & 0.359 & 0.566 & 0.823 & 0.017 \\
BPE Claude (all-eq.) & 4.24 {\scriptsize $\pm$12.28} & 4.6684 & 0.142 & 0.360 & 0.531 & 0.811 & 0.016 \\
BPE Punct (all-eq.) & 4.62 {\scriptsize $\pm$12.21} & 4.2193 & 0.091 & 0.383 & 0.525 & 0.765 & 0.015 \\
BPE GPT-4o NFC (all-eq.) & 4.18 {\scriptsize $\pm$12.10} & 4.7119 & 0.140 & 0.359 & 0.566 & 0.823 & 0.017 \\
BPE GPT-4o (all-eq., no-rep.) & 4.18 {\scriptsize $\pm$12.10} & 4.7119 & 0.140 & 0.359 & 0.566 & 0.823 & 0.017 \\
BPE GPT-4o (bal., mf50) & 4.36 {\scriptsize $\pm$11.29} & 4.2847 & 0.132 & 0.407 & 0.578 & 0.801 & 0.033 \\
BPE NFC clean (bal.) & 4.42 {\scriptsize $\pm$11.18} & 4.2056 & 0.113 & 0.413 & 0.541 & 0.783 & 0.033 \\
BPE NFC clean plus2 (bal.) & 4.47 {\scriptsize $\pm$11.64} & 4.2161 & 0.132 & 0.425 & 0.541 & 0.783 & 0.031 \\
BPE NFC clean plus3 (bal.) & 4.47 {\scriptsize $\pm$11.65} & 4.2127 & 0.132 & 0.425 & 0.541 & 0.783 & 0.031 \\
SuperBPE PA-BPE GPT-4o & 4.16 {\scriptsize $\pm$12.35} & 4.9042 & 0.145 & 0.330 & 0.642 & 0.846 & 0.019 \\
SuperBPE PA-BPE GPT-4o t64k & 4.14 {\scriptsize $\pm$11.92} & 4.8166 & 0.143 & 0.383 & 0.657 & 0.844 & 0.020 \\
SuperBPE clean C2 (bal.) & 4.31 {\scriptsize $\pm$12.51} & 4.6590 & 0.145 & 0.373 & 0.529 & 0.809 & 0.018 \\
SuperBPE clean C3 (bal.) & 4.19 {\scriptsize $\pm$12.27} & 4.8310 & 0.144 & 0.345 & 0.592 & 0.835 & 0.019 \\
SuperBPE Punct t64k & 4.46 {\scriptsize $\pm$11.38} & 4.2271 & 0.120 & 0.443 & 0.626 & 0.810 & 0.034 \\
PA-BPE NFC GPT-4o HW & 4.12 {\scriptsize $\pm$12.01} & 4.8303 & 0.140 & 0.287 & 0.563 & 0.828 & 0.016 \\
PA-BPE NFC clean & 4.38 {\scriptsize $\pm$12.98} & 4.6207 & 0.146 & 0.273 & 0.527 & 0.801 & 0.018 \\
Unigram GPT-4o tuned & 4.40 {\scriptsize $\pm$9.21} & 3.5078 & 0.171 & 0.374 & 0.329 & 0.690 & 0.045 \\
BPE GPT-4o (highres) & 6.49 {\scriptsize $\pm$14.16} & 2.7379 & 0.183 & 0.483 & 0.505 & 0.677 & 0.227 \\
BPE GPT-4o (highmid) & 5.07 {\scriptsize $\pm$11.34} & 3.4759 & 0.207 & 0.417 & 0.504 & 0.734 & 0.158 \\
BPE Whitespace (bal.) & 4.41 {\scriptsize $\pm$11.53} & 4.2929 & 0.130 & 0.406 & 0.696 & 0.815 & 0.037 \\
\bottomrule
\end{tabular}}
 \end{table*}

\section{Language Model Results}\label{app:additional-results}

\subsection{Language Model Evaluation Results (Full-Scale; 1.27B)}\label{app:lm_eval_results}

\cref{tab:main-results} reports absolute scores per model on downstream benchmarks for every tokenizer named in this paper, which is a superset of \primarypanel{}: it adds the extended-panel-only variants, the two defective English-only retrains, and several controls used in the appendix analyses. \cref{tab:tokenizer-grid} gives each row's panel membership and configuration.

\begin{table*}[t]
\centering
\caption{Evaluation results for the 1.27B models: \nStdAll{} tokenizers trained on the standard corpus mixture at a $\sim$128K vocabulary, plus two off-the-shelf reference tokenizers (marked with $^\dagger$). Panel membership per row is in \cref{tab:tokenizer-grid}. Lower is better for the BPB columns; higher is better for BLiMP and MultiBLiMP. Code BPB for the 25 tokenizers in the cross-model comparison appears in \cref{tab:cross-regime-code-bpb}, where the two model families' ranks, not absolute scores, are compared. `--': not evaluated for that tokenizer.}
\label{tab:main-results}
\adjustbox{max width=\linewidth}{
\small
\begin{tabular}{l c cc cc cc}
\toprule
 & \multicolumn{5}{c}{\textbf{Perplexity $\downarrow$} (mean $|$ cross-language $\sigma$)} & \multicolumn{2}{c}{\textbf{Ling.} $\uparrow$} \\
\cmidrule(lr){2-6} \cmidrule(lr){7-8}
\textbf{Tokenizer} & \textbf{Val} & \textbf{FLORES (tr.)} & \textbf{$\sigma$} & \textbf{FLORES (all)} & \textbf{$\sigma$} & \textbf{BLiMP} & \textbf{MultiBLiMP} \\
\midrule
Mistral-Nemo$^\dagger$ & 0.7212 & 1.173 & 0.299 & 2.756 & 1.012 & 0.817 & 0.915 \\
LLaMA-3$^\dagger$ & 0.7198 & 1.173 & 0.293 & 2.662 & 1.014 & 0.826 & 0.914 \\
\midrule
BPE Punct (bal.) & 0.7185 & 1.167 & 0.291 & 2.647 & 1.018 & 0.817 & 0.913 \\
BPE Punct (eng.) & 0.7360 & 1.191 & 0.308 & 2.636 & 1.015 & 0.807 & 0.911 \\
BPE GPT-4o (bal.) & 0.7131 & 1.161 & 0.294 & 2.641 & 1.020 & 0.820 & 0.916 \\
BPE GPT-4o NFC & 0.7133 & 1.159 & 0.298 & 2.638 & 1.019 & 0.820 & 0.920 \\
BPE GPT-4o (eng.) & 0.7375 & 1.197 & 0.312 & 2.638 & 1.015 & 0.807 & 0.911 \\
BPE GPT-4o (eng., full-byte) & 0.7382 & 1.201 & 0.316 & 2.644 & 1.017 & 0.817 & 0.912 \\
BPE GPT-4o (code) & 0.7252 & 1.179 & 0.295 & 2.626 & 1.009 & 0.819 & 0.913 \\
BPE Claude (bal.) & 0.7156 & 1.162 & 0.296 & 2.642 & 1.016 & 0.816 & 0.914 \\
BPE Claude NFC & 0.7152 & 1.159 & 0.300 & 2.642 & 1.018 & 0.819 & 0.911 \\
BPE Claude (eng.) & 0.7330 & 1.194 & 0.311 & 2.641 & 1.021 & 0.817 & 0.907 \\
BPE RightAlign (bal.) & 0.7138 & 1.164 & 0.293 & 2.641 & 1.016 & 0.813 & 0.909 \\
BPE RightAlign NFC & 0.7138 & 1.161 & 0.298 & 2.642 & 1.018 & 0.816 & 0.911 \\
Unigram GPT-4o & 0.7338 & 1.193 & 0.304 & 2.687 & 1.010 & 0.834 & 0.903 \\
Unigram Claude & 0.7359 & 1.189 & 0.302 & 2.683 & 1.007 & 0.841 & 0.905 \\
Unigram RightAlign & 0.7338 & 1.193 & 0.302 & 2.684 & 1.002 & 0.828 & 0.908 \\
SuperBPE GPT-4o (bal.) & 0.7236 & 1.173 & 0.297 & 2.639 & 1.013 & 0.804 & 0.909 \\
PA-BPE NFC GPT-4o & 0.7255 & 1.186 & 0.300 & 2.662 & 1.006 & 0.817 & 0.904 \\
Unigram GPT-4o (highmid) & 0.7345 & 1.197 & 0.306 & 2.675 & 1.003 & 0.831 & 0.913 \\
Unigram GPT-4o (highres) & 0.7305 & 1.197 & 0.305 & 2.666 & 0.994 & 0.824 & 0.907 \\
MinGram SCRIPT-enc & 0.7129 & 1.165 & 0.303 & 2.685 & 1.048 & 0.813 & 0.920 \\
MinGram SCRIPT-enc (nl-split) & 0.7134 & 1.159 & 0.301 & 2.671 & 1.058 & 0.815 & 0.918 \\
BPE SCRIPT-enc GPT-4o & 0.7128 & 1.157 & 0.297 & 2.657 & 1.046 & 0.822 & 0.916 \\
BPE GPT-2 (bal.) & 0.7128 & 1.157 & 0.287 & 2.635 & 1.016 & 0.816 & 0.910 \\
BPE GPT-4o (all-eq.) & 0.7177 & 1.172 & 0.292 & 2.652 & 1.016 & 0.819 & 0.918 \\
BPE Claude (all-eq.) & 0.7200 & 1.170 & 0.291 & 2.654 & 1.015 & 0.817 & 0.921 \\
BPE Punct (all-eq.) & 0.7223 & 1.168 & 0.294 & 2.648 & 1.019 & 0.818 & 0.917 \\
BPE GPT-4o NFC (all-eq.) & 0.7165 & 1.164 & 0.295 & 2.646 & 1.014 & 0.823 & 0.917 \\
BPE GPT-4o (all-eq., no-rep.) & 0.7165 & 1.168 & 0.292 & 2.648 & 1.015 & 0.819 & 0.913 \\
BPE GPT-4o (bal., mf50) & 0.7116 & 1.158 & 0.292 & 2.629 & 1.013 & 0.822 & -- \\
BPE NFC clean (bal.) & 0.7159 & 1.157 & 0.296 & 2.648 & 1.022 & 0.821 & 0.910 \\
BPE NFC clean plus2 (bal.) & 0.7162 & 1.158 & 0.297 & 2.638 & 1.017 & 0.814 & 0.918 \\
BPE NFC clean plus3 (bal.) & 0.7149 & 1.155 & 0.297 & 2.635 & 1.017 & 0.818 & 0.918 \\
SuperBPE PA-BPE GPT-4o & 0.7292 & 1.181 & 0.294 & 2.653 & 1.008 & 0.801 & 0.916 \\
SuperBPE PA-BPE GPT-4o t64k & 0.7293 & 1.181 & 0.295 & 2.652 & 1.011 & 0.792 & 0.920 \\
SuperBPE clean C2 (bal.) & 0.7285 & 1.169 & 0.293 & 2.650 & 1.011 & 0.811 & 0.912 \\
SuperBPE clean C3 (bal.) & 0.7296 & 1.174 & 0.294 & 2.644 & 1.006 & 0.803 & 0.919 \\
SuperBPE Punct t64k & 0.7259 & 1.167 & 0.293 & 2.630 & 1.009 & 0.808 & 0.915 \\
PA-BPE NFC GPT-4o HW & 0.7194 & 1.177 & 0.299 & 2.656 & 1.007 & 0.816 & 0.914 \\
PA-BPE NFC clean & 0.7235 & 1.178 & 0.302 & 2.653 & 1.007 & 0.819 & 0.907 \\
Unigram GPT-4o tuned & 0.7333 & 1.192 & 0.302 & 2.686 & 1.009 & 0.838 & 0.912 \\
BPE GPT-4o (highres) & 0.7127 & 1.182 & 0.294 & 2.618 & 0.999 & 0.814 & 0.912 \\
BPE GPT-4o (highmid) & 0.7137 & 1.174 & 0.296 & 2.637 & 1.016 & 0.815 & 0.910 \\
BPE Whitespace (bal.) & 0.7207 & 1.165 & 0.294 & 2.622 & 1.003 & 0.814 & 0.915 \\
Unigram Whitespace tuned (bal.) & 0.8401 & 1.176 & 0.298 & 2.576 & 0.949 & 0.805 & 0.918 \\
\bottomrule
\end{tabular}
}
\end{table*}

\subsection{Language Model Evaluation Results (Smaller-Model-Scale; $\sim$596M)}
\label{app:pilot}

We also train $\sim$596M-parameter models (nanochat d16: 16 layers, 1024 hidden dimension, 8 heads; the parameter total is counted as for the 1.27B models, i.e., including the input, output, and value embeddings, which at this width are $394$M of the total) on the same data; \cref{tab:pilot-results} reports val BPB, FLORES BPB, and code BPB for this scale. Note that these models cover only a subset of tokenizers from the entire panel, in particular, the ones that were part of the initial panel. 
These use 3.5B tokens following the same 10.5$\times$ scaling ratio from the nanochat library.
Smaller models models serve as a fast iteration loop; between the 596M and 1.27B scales the tokenizer ranking by validation BPB has Kendall $\tau = 0.752$ (\cref{app:cross-scale-ranking}).

\begin{table}[h]
\centering
\caption{Smaller model ($\sim$596M) evaluation results; lower is better for all metrics. $^\dagger$: off-the-shelf reference tokenizer, as in \cref{tab:main-results}. The roster is the 30 tokenizers of \cref{tab:main-results} that also have a model at this scale, plus the same 2 references; the remaining configurations were added after the smaller-model sweep and have no 596M model.}
\label{tab:pilot-results}
\small
\begin{tabular}{lcccc}
\toprule
\textbf{Tokenizer} & \textbf{Val BPB} & \textbf{FLORES (tr.)} & \textbf{FLORES (all)} & \textbf{Code BPB} \\
\midrule
Mistral-Nemo$^\dagger$ & 0.8210 & 1.292 & 2.827 & 0.612 \\
LLaMA-3$^\dagger$ & 0.8193 & 1.304 & 2.751 & 0.622 \\
\midrule
BPE Punct (bal.) & 0.8179 & 1.295 & 2.737 & 0.613 \\
BPE Punct (eng.) & 0.8334 & 1.343 & 2.748 & 0.646 \\
BPE GPT-4o (bal.) & 0.8067 & 1.290 & 2.734 & 0.599 \\
BPE GPT-4o NFC & 0.8064 & 1.287 & 2.740 & 0.599 \\
BPE GPT-4o (eng.) & 0.8327 & 1.344 & 2.749 & 0.648 \\
BPE GPT-4o (code) & 0.8282 & 1.332 & 2.737 & 0.625 \\
BPE Claude (bal.) & 0.8131 & 1.291 & 2.740 & 0.599 \\
BPE Claude NFC & 0.8138 & 1.288 & 2.736 & 0.602 \\
BPE Claude (eng.) & 0.8327 & 1.343 & 2.748 & 0.639 \\
BPE RightAlign (bal.) & 0.8066 & 1.289 & 2.736 & 0.599 \\
BPE RightAlign NFC & 0.8065 & 1.286 & 2.734 & 0.599 \\
Unigram GPT-4o & 0.8407 & 1.328 & 2.786 & 0.642 \\
Unigram Claude & 0.8471 & 1.334 & 2.791 & 0.643 \\
Unigram RightAlign & 0.8408 & 1.326 & 2.790 & 0.642 \\
PA-BPE NFC GPT-4o & 0.8333 & 1.299 & 2.760 & 0.624 \\
PA-BPE NFC clean & 0.8309 & 1.294 & 2.752 & 0.614 \\
BPE Whitespace (bal.) & 0.8101 & 1.297 & 2.725 & 0.603 \\
BPE GPT-2 (bal.) & 0.8083 & 1.292 & 2.730 & 0.596 \\
BPE GPT-4o (highres) & 0.8084 & 1.323 & 2.721 & 0.603 \\
BPE GPT-4o (highmid) & 0.8053 & 1.300 & 2.728 & 0.598 \\
Unigram GPT-4o tuned & 0.8440 & 1.335 & 2.801 & 0.644 \\
Unigram Whitespace tuned (bal.) & 0.9633 & 1.312 & 2.692 & 0.720 \\
PA-BPE NFC GPT-4o HW & 0.8165 & 1.280 & 2.737 & 0.606 \\
SuperBPE Punct t64k & 0.8127 & 1.298 & 2.720 & 0.594 \\
SuperBPE GPT-4o (bal.) & 0.8098 & 1.302 & 2.729 & 0.592 \\
SuperBPE PA-BPE GPT-4o & 0.8214 & 1.291 & 2.742 & 0.607 \\
SuperBPE PA-BPE GPT-4o t64k & 0.8191 & 1.293 & 2.742 & 0.600 \\
SuperBPE clean C2 (bal.) & 0.8262 & 1.279 & 2.734 & 0.607 \\
SuperBPE clean C3 (bal.) & 0.8214 & 1.283 & 2.737 & 0.609 \\
BPE NFC clean (bal.) & 0.8165 & 1.286 & 2.740 & 0.605 \\
\bottomrule
\end{tabular}
 \end{table}

\subsection{Ranking Stability Across Training Scales}\label{app:cross-scale-ranking}
We report Kendall's $\tau$ between per-tokenizer val BPB rank vectors at each pair of the
four training scales (d8 $\sim$222M, d12 $\sim$381M, d16 $\sim$596M, d24 1.27B), for the
18-tokenizer panel, in \cref{tab:cross-scale-ranking}. All models at each scale were trained
under one software configuration for that scale.
Adjacent-scale $\tau$ is $0.882$ (d8-d12), $0.922$ (d12-d16), and $0.752$ (d16-d24); the
full-versus-smallest (d8-d24) $\tau$ is $0.765$. 
Notably, the d16-d24 pair has the lowest $\tau$ of
the three adjacent pairs, suggesting the danger of using results from smaller language models for larger language model tokenizer development. The ranking changes most at the final doubling of scale, from 596M to the 1.27B target: $\tau = 0.752$ corresponds to about one tokenizer pair in eight ordered differently at the two scales. Rank agreement between adjacent scales does not tighten as scale grows within our range, so a tokenizer selected on val BPB at a smaller scale can sit several rank positions away at the target scale, and a small-scale sweep is a screening tool rather than a substitute for a target-scale comparison.

\subsection{Cross-Model Rank Correlation of Code BPB}\label{app:cross-regime-code-bpb}

While the math+code models (\cref{app:mc20b-results}) share tokenizers with the natural-language-focused models, they are trained on very different data mixtures, so code BPB levels are not comparable between the two by construction: for a given tokenizer, a lower value under a math+code model does not mean we should expect lower code BPB under the corresponding natural-language-focused model. \cref{tab:cross-regime-code-bpb} compares code BPB ranks between the two model families: because rank transfer is a per-tokenizer comparison rather than a correlation panel, this set keeps every eligible tokenizer, including family variants that the primary panel collapses to one representative, both English-only GPT-4o variants (with and without the full byte alphabet, \cref{app:tok-inclusion}), and the whitespace minimal-structure probe.
The rank correlation between the natural-language models' code BPB rank and the math+code models' code BPB rank is Spearman $\rho = 0.307$ (95\% bootstrap CI $-0.191$ to $0.715$, $n=25$, with the labeled control included) and $\rho = 0.359$ (95\% CI $-0.147$ to $0.764$, $n=24$) with it excluded. Both confidence intervals span zero, so the natural-language models' code BPB ranking does not transfer detectably to the math+code models' code BPB ranking. We took this result as further justification of the choice to train dedicated math and code models.

\begin{table}[t]
\centering
\caption{Code BPB and code BPB rank for the natural-language models and the math+code models, for \crnew{the 25 tokenizers with a model in both families}, sorted by natural-language-model rank. Code BPB levels are not comparable across the two columns; only the rank columns are.
$^\dagger$: labeled control \crnew{(whitespace minimal-structure probe)}, excluded from the without-control correlation reported in the text.
\crnew{$^{*}$: this row's math+code model was trained before a change to the cluster software environment that the other 24 postdate. It is kept because it is the only math+code run for that tokenizer.} \crnew{The Code BPB caveat of \cref{app:tok-inclusion} applies to the three English-only rows without a full byte alphabet.}}
\label{tab:cross-regime-code-bpb}
\adjustbox{max width=\linewidth}{
\small
\begin{tabular}{lcccc}
\toprule
\textbf{Tokenizer} & \textbf{\crnew{Natural-language} BPB} & \textbf{\crnew{Natural-language} rank} & \textbf{Math+code BPB} & \textbf{Math+code rank} \\
\midrule
MinGram SCRIPT-enc & $0.5125$ & $1.0$ & $0.3793$ & $9.0$ \\
BPE SCRIPT-enc GPT-4o & $0.5186$ & $2.0$ & $0.3781$ & $6.0$ \\
BPE Claude (bal.) & $0.5203$ & $3.0$ & $0.3861$ & $18.0$ \\
BPE GPT-4o (bal.)$^{*}$ & $0.5206$ & $4.0$ & $0.3773$ & $5.0$ \\
BPE Claude NFC & $0.5206$ & $5.0$ & $0.3845$ & $17.0$ \\
BPE GPT-4o NFC & $0.5207$ & $6.0$ & $0.3786$ & $8.0$ \\
BPE RightAlign (bal.) & $0.5223$ & $7.0$ & $0.3798$ & $10.0$ \\
BPE RightAlign NFC & $0.5225$ & $8.0$ & $0.3809$ & $13.0$ \\
BPE NFC clean (bal.) & $0.5226$ & $9.0$ & $0.3815$ & $14.0$ \\
BPE NFC clean plus3 (bal.) & $0.5232$ & $10.0$ & $0.3834$ & $16.0$ \\
BPE NFC clean plus2 (bal.) & $0.5249$ & $11.0$ & $0.3785$ & $7.0$ \\
SuperBPE GPT-4o (bal.) & $0.5249$ & $12.0$ & $0.3834$ & $15.0$ \\
SuperBPE clean C2 (bal.) & $0.5259$ & $13.0$ & $0.3906$ & $21.0$ \\
BPE Whitespace (bal.)$^\dagger$ & $0.5269$ & $14.0$ & $0.3742$ & $4.0$ \\
BPE Punct (bal.) & $0.5302$ & $15.0$ & $0.3801$ & $11.0$ \\
SuperBPE clean C3 (bal.) & $0.5306$ & $16.0$ & $0.3896$ & $20.0$ \\
PA-BPE NFC GPT-4o & $0.5335$ & $17.0$ & $0.3878$ & $19.0$ \\
BPE Claude (eng.) & $0.5426$ & $18.0$ & $0.3652$ & $1.0$ \\
BPE GPT-4o (code) & $0.5447$ & $19.0$ & $0.3805$ & $12.0$ \\
BPE Punct (eng.) & $0.5551$ & $20.0$ & $0.3724$ & $3.0$ \\
Unigram RightAlign & $0.5566$ & $21.0$ & $0.3989$ & $22.0$ \\
BPE GPT-4o (eng.) & $0.5567$ & $22.0$ & $0.3709$ & $2.0$ \\
BPE GPT-4o (eng., full-byte) & $0.5568$ & $23.0$ & $0.3991$ & $23.0$ \\
Unigram GPT-4o & $0.5569$ & $24.0$ & $0.3993$ & $24.0$ \\
Unigram Claude & $0.5578$ & $25.0$ & $0.4032$ & $25.0$ \\
\bottomrule
\end{tabular}
}
\end{table}

\section{Additional Analyses}\label{app:additional-analyses}

\subsection{Held-Out Prediction of the Reference Tokenizers}\label{app:heldout-reference-prediction}

We test how far a fit estimated on the trained tokenizer panel predicts a tokenizer the panel never saw. For each of four targets, val BPB and FLORES trained-31 BPB (fit on the \nExtended{}-tokenizer extended panel of \cref{sec:tokenizer-configs,app:tok-inclusion}), and code BPB and MBPP pass@1 from the math+code models (both fit on the \nMCPrimary{}-member math+code panel, \mathcodepanel{}; \cref{tab:panel-overview}), we take the best univariate intrinsic predictor and a multivariate fit on up to 3 standardized predictors selected by leave-one-family-out cross-validation inside the panel (tokenizers from the same family, i.e., same algorithm, general pretokenization strategy, and training data, are held out together, so no fold trains on a tokenizer from its test point's family), then predict the two off-the-shelf reference tokenizers' models, Mistral-Nemo and LLaMA-3, which share architecture and training data with the panel but were excluded from every panel fit. We report each reference's prediction error standardized by the fit's cross-validated root-mean-squared error (CV-RMSE), and whether the observed value falls inside the CV-RMSE-calibrated 90\% prediction interval (constructed as described in \cref{app:stat-protocol}).

\begin{table}[t]
\centering
\caption{\crnew{Held-out prediction of the two off-the-shelf reference tokenizers. For each downstream target, the best univariate intrinsic predictor and the multivariate fit (up to 3 standardized predictors, selected by leave-one-family-out cross-validation), with the in-sample $R^2$, the cross-validated root-mean-squared error (CV-RMSE), and each reference's prediction error divided by the CV-RMSE (standardized error). $^{\dagger}$: the observed value falls outside the fit's 90\% prediction interval. Intrinsic predictors are measured on FLORES+ unless marked otherwise. Val BPB and FLORES (trained) BPB are fit on the \nExtended{}-tokenizer extended panel (\extendedpanel); the two math+code targets are fit on the \nMCPrimary{}-tokenizer math+code panel.}}
\label{tab:heldout-reference}
\adjustbox{max width=\linewidth}{
\small
\setlength{\tabcolsep}{4pt}
\begin{tabular}{@{}p{2.3cm}l p{3.5cm} c c r r@{}}
\toprule
 & & & & & \multicolumn{2}{c}{Standardized error} \\
\cmidrule(lr){6-7}
Target & Fit & Predictors & $R^2$ & CV-RMSE & \shortstack[r]{Mistral-\\Nemo} & LLaMA-3 \\
\midrule
Val BPB ($n{=}39$) & univ. & tokens per identifier & 0.53 & 0.0061 & $+0.48$ & $+1.43$ \\
 & multiv. & fertility, Gini coefficient, tokens per identifier & 0.68 & 0.0054 & $+0.35$ & $+1.84^{\dagger}$ \\
\addlinespace
FLORES (trained) BPB ($n{=}39$) & univ. & identifier fragmentation & 0.59 & 0.0091 & $+0.36$ & $+2.10^{\dagger}$ \\
 & multiv. & fertility, identifier fragmentation, indentation consistency & 0.81 & 0.0062 & $+0.41$ & $+2.18^{\dagger}$ \\
\addlinespace
Code BPB (math+code) ($n{=}20$) & univ. & Gini coefficient (FineWeb) & 0.48 & 0.0074 & $-0.40$ & $-2.01^{\dagger}$ \\
 & multiv. & vocabulary utilization (FineWeb), trigram entropy, operator isolation (code corpus) & 0.68 & 0.0070 & $-0.05$ & $-0.90$ \\
\addlinespace
MBPP pass@1 (math+code) ($n{=}20$) & univ. & AST boundary alignment & 0.26 & 0.0793 & $-1.15$ & $-0.03$ \\
 & multiv. & token length, unigram entropy, AST boundary alignment & 0.50 & 0.0739 & $-1.78^{\dagger}$ & $+1.77^{\dagger}$ \\
\bottomrule
\end{tabular}
}
\end{table}

\cref{tab:heldout-reference} reports all eight fits. Under the best univariate fits, 6 of the 8 reference-target predictions (4 targets times 2 references) fall inside their 90\% prediction interval; both exceptions are LLaMA-3, on FLORES (trained) BPB at $+2.10$ standardized errors and on the math+code models' code BPB at $-2.01$. The multivariate fits have a higher in-sample $R^2$ and a lower CV-RMSE than the univariate fits on every target, yet 4 of their 8 reference predictions fall outside the 90\% interval, against 2 of 8 for the univariate fits: the narrower multivariate intervals overstate the precision available for a tokenizer from outside the panel.

The selected predictors differ across the two sets of language models we train (which likewise differ in the downstream tasks on which they're evaluated). The natural-language models' val BPB and FLORES (trained) BPB are best predicted by tokens per identifier and identifier fragmentation; the math+code models' code BPB is best predicted by the Gini coefficient measured on FineWeb, and for the univariate regression models, none of its selected predictors is an entropy metric (unigram, bigram, or trigram entropy, or R\'enyi efficiency).

\subsection{Intrinsic--Downstream Correlations Robustness Checks}\label{app:robustness}

\cref{tab:aggregate-correlations-fineweb} repeats the aggregate correlation analysis of \cref{tab:aggregate-correlations}, with the FLORES+-measured intrinsic metrics replaced by their FineWeb-Edu/FineWeb2-measured counterparts, over the same primary panel and downstream targets. Four rows (identifier fragmentation, digit boundary $F_1$, operator isolation on the math corpus, and AST alignment) are byte-identical between the two tables, since those metrics are computed on code and math corpora rather than on the natural-language corpus that varies between them; see the caption below.

\begin{table*}[t]
\centering
\caption{Aggregate Spearman (FineWeb-2/-Edu intrinsic) $\rho$ between intrinsic metrics and downstream
performance \crnew{($n=\nPrimary{}$ primary-panel tokenizers; $n=\nMCInStdPanel{}$ for the math+code Code BPB and MBPP columns, the primary-panel members that also have a math+code model)}.
Intrinsic metrics are measured on FineWeb-Edu and FineWeb2 with length unit $u=\texttt{bytes}$, so
unit-dependent metrics (compression rate, fertility) are not comparable with the same column name in
\cref{tab:aggregate-correlations}, which uses $u=\texttt{lines}$.
Four rows, Ident.\ fragmentation, Digit boundary F1, Operator isolation (math), and AST alignment, are computed on code and math corpora rather than on FineWeb, so they are byte-identical to the same rows of \cref{tab:aggregate-correlations}.
\crnew{\textbf{Bold} indicates significance, with stars marking the (BH-adjusted) thresholds, over the 98 printed cells and exactly as in \cref{tab:aggregate-correlations}: $^{*}$\,$p_{\text{adj}}<0.05$, $^{**}$\,$p_{\text{adj}}<0.01$, $^{***}$\,$p_{\text{adj}}<0.001$.}}
\label{tab:aggregate-correlations-fineweb}
\adjustbox{max width=\linewidth}{
\small
\begin{tabular}{lccccccc}
\toprule
\multirow{2}{*}{\textbf{Intrinsic Metric}} & \multirow{2}{*}{\textbf{Val BPB}} & \multicolumn{2}{c}{\textbf{FLORES}} & \multicolumn{2}{c}{\textbf{Math+code}} & \multirow{2}{*}{\textbf{BLiMP}} & \multirow{2}{*}{\textbf{MultiBLiMP}} \\
\cmidrule(lr){3-4} \cmidrule(lr){5-6}
 &  & \textbf{trained} & \textbf{all} & \textbf{Code BPB} & \textbf{MBPP} &  &  \\
\midrule
Digit boundary F1 & $-0.42$ & $-0.41$ & $\textbf{-0.67}^{**}$ & $\textbf{-0.62}^{*}$ & $0.21$ & $\textbf{-0.51}^{*}$ & $0.32$ \\
Ident.\ fragmentation & $\textbf{0.53}^{*}$ & $\textbf{0.57}^{*}$ & $0.36$ & $0.41$ & $-0.21$ & $0.37$ & $-0.45$ \\
R\'enyi eff.\ ($\alpha$=2) & $\textbf{-0.53}^{*}$ & $\textbf{-0.72}^{***}$ & $-0.42$ & $-0.49$ & $0.12$ & $-0.25$ & $0.32$ \\
Operator isolation (math) & $\textbf{0.54}^{*}$ & $0.38$ & $\textbf{0.55}^{*}$ & $0.39$ & $0.27$ & $0.32$ & $-0.25$ \\
Gini coefficient & $0.33$ & $\textbf{0.62}^{**}$ & $0.38$ & $\textbf{0.60}^{*}$ & $-0.10$ & $0.24$ & $-0.36$ \\
UTF-8 char split & $0.38$ & $\textbf{0.56}^{*}$ & $-0.33$ & $0.25$ & $-0.26$ & $0.09$ & $-0.43$ \\
Vocab utilization & $0.26$ & $0.07$ & $\textbf{0.70}^{**}$ & $0.54$ & $0.05$ & $0.32$ & $-0.17$ \\
Trigram entropy & $-0.37$ & $\textbf{-0.54}^{*}$ & $0.12$ & $-0.28$ & $0.08$ & $-0.14$ & $0.33$ \\
UTF-8 boundary crossing & $0.18$ & $0.34$ & $-0.33$ & $0.20$ & $-0.27$ & $0.18$ & $-0.33$ \\
Fertility & $0.27$ & $0.41$ & $-0.37$ & $0.16$ & $0.23$ & $-0.04$ & $-0.28$ \\
Compression rate & $-0.32$ & $-0.47$ & $0.24$ & $-0.19$ & $-0.03$ & $-0.17$ & $0.28$ \\
Bigram entropy & $-0.35$ & $\textbf{-0.53}^{*}$ & $0.00$ & $-0.21$ & $-0.02$ & $-0.23$ & $0.30$ \\
AST alignment & $0.31$ & $0.01$ & $0.36$ & $0.16$ & $\textbf{0.61}^{*}$ & $0.10$ & $-0.05$ \\
Unigram entropy & $-0.34$ & $\textbf{-0.51}^{*}$ & $0.04$ & $-0.12$ & $0.04$ & $-0.22$ & $0.26$ \\
\bottomrule
\end{tabular}
}
\end{table*}

As a robustness check, we repeat the correlation analysis at the extended-panel level (\cref{sec:evaluation}).\footnote{Each of the \nFamilies{} tokenizer families in the \nExtended{}-tokenizer extended panel is collapsed to its family-mean score before computing the Spearman correlation, so \crnew{same-family} tokenizers cannot count as independent data points, and significance is assessed by a permutation test that reshuffles whole families rather than individual tokenizers. 
This extended-level correction is computed over a larger set of comparisons than the 98 intrinsic metric--downstream performance correlations printed in \cref{tab:aggregate-correlations}, so its adjusted $p$-values are not directly comparable to the table's.} Under this check, the R\'enyi-efficiency and digit-boundary-$F_1$ correlations reported above both remain significant ($p_{\text{adj}} = 0.016$ and $0.029$), as do four of the five information-theoretic correlations with FLORES(trained) BPB; the fifth, unigram entropy, falls just short of significance ($p_{\text{adj}} = 0.062$).
\crnew{We checked whether any single tokenizer drives the four correlations discussed above (R\'enyi efficiency with FLORES(trained) BPB, digit boundary $F_1$ with BLiMP, digit boundary $F_1$ with Code BPB, and AST alignment with MBPP) by removing tokenizers one at a time from the relevant panel and recomputing each correlation. Every one of these leave-one-out recomputations keeps the original sign and $p<0.05$ before correction for multiple testing: this holds across all \nPrimary{} possible single-tokenizer removals for the two correlations computed on the full primary panel (R\'enyi efficiency/FLORES and digit boundary $F_1$/BLiMP), and across all \nMCInStdPanel{} possible removals for the two correlations restricted to primary-panel tokenizers with a math+code model (digit boundary $F_1$/Code BPB and AST alignment/MBPP); the worst-case $p$-value across every removal of all four correlations is $0.0162$. Under the correction applied jointly across the full table (\cref{tab:aggregate-correlations}), removing a single tokenizer, Unigram Claude (Claude pretokenizer, UnigramLM algorithm), is enough to push three of these four correlations, digit boundary $F_1$ with BLiMP, digit boundary $F_1$ with Code BPB and AST alignment with MBPP, just above the corrected significance threshold, and 12 of the 29 possible single removals move at least one of the four.}

\crnew{We also checked robustness to removing whole groups of tokenizers that share a non-default value on one design axis (algorithm, pretokenization, or training-data composition; \cref{sec:tokenizer-configs}), keeping only the tokenizers that share this panel's single most common value on that axis. These axis-based removals affect the four correlations less uniformly than single-tokenizer removal: three weaken but keep their sign, while the AST-alignment/MBPP correlation reverses sign, becoming small and not significant ($\rho = -0.25$), when restricted to the 7 tokenizers in this panel that use the GPT-4o pretokenizer, this panel's most common pretokenization choice. This reversal is consistent with the line-break attribution below: the AST-alignment/MBPP association reflects a difference between pretokenizer choices rather than a relationship that holds within a single pretokenizer choice.}

\begin{table}[t]
\centering
\caption{Kendall's $\tau$ between per-tokenizer val BPB rank vectors across the four training
scales, 18-tokenizer panel ($n=18$ for every cell; unlike the other tables in this paper, this
panel includes the two off-the-shelf reference tokenizers). All models at each scale were trained under
one software configuration for that scale.}
\label{tab:cross-scale-ranking}
\begin{tabular}{lcccc}
\toprule
 & \textbf{d8 (tiny)} & \textbf{d12 (small)} & \textbf{d16 (pilot)} & \textbf{d24 (full)} \\
\midrule
\textbf{d8 (tiny)} & $1.000$ & $0.882$ & $0.882$ & $0.765$ \\
\textbf{d12 (small)} & $0.882$ & $1.000$ & $0.922$ & $0.725$ \\
\textbf{d16 (pilot)} & $0.882$ & $0.922$ & $1.000$ & $0.752$ \\
\textbf{d24 (full)} & $0.765$ & $0.725$ & $0.752$ & $1.000$ \\
\bottomrule
\end{tabular}
 \end{table}

\subsection{Math and Code Generation (math+code Models)}\label{app:mc20b-results}\label{app:newline-mechanism}
\begin{table*}[t]
\centering
\caption{Evaluation results for the math+code models: \crnew{\nMCExtended{}} LMs trained from
scratch on a math and code mixture (\cref{sec:training-details}) using custom tokenizers at a $\sim$128K vocabulary, plus two off-the-shelf
reference tokenizers (marked with $^\dagger$). \crnew{$^{b}$: this row's tokenizer cannot represent all 256 byte values; \cref{app:tok-inclusion} gives the inclusion rule for the two marked rows.} \crnew{$^{e}$: a member of the extended math+code roster only, not one of the \nMCPrimary{} tokenizers of the math+code correlation panel (\mathcodepanel{}); the roster behind each correlation in \cref{app:mc20b-results} is stated where that correlation is reported.} Lower is better for Code BPB; higher is better for MBPP, HumanEval, and GSM8K.}
\label{tab:mc20b-results}
\small
\begin{tabular}{lcccc}
\toprule
\textbf{Tokenizer} & \textbf{MBPP} & \textbf{HumanEval} & \textbf{GSM8K} & \textbf{Code BPB} \\
\midrule
BPE Punct (bal.) & 0.230 & 0.171 & 0.189 & 0.380 \\
BPE Punct (eng.)$^{b}$ & 0.228 & 0.165 & 0.208 & 0.372 \\
BPE GPT-4o (bal.) & 0.174 & 0.165 & 0.198 & 0.377 \\
BPE GPT-4o NFC & 0.120 & 0.152 & 0.208 & 0.379 \\
BPE GPT-4o (eng., full-byte) & 0.064 & 0.152 & 0.218 & 0.399 \\
BPE GPT-4o (code) & 0.016 & 0.159 & 0.216 & 0.380 \\
BPE Claude (bal.) & 0.250 & 0.171 & 0.227 & 0.386 \\
BPE Claude NFC & 0.224 & 0.159 & 0.230 & 0.385 \\
BPE Claude (eng.)$^{b}$ & 0.234 & 0.165 & 0.221 & 0.365 \\
BPE RightAlign (bal.) & 0.072 & 0.152 & 0.234 & 0.380 \\
BPE RightAlign NFC & 0.104 & 0.171 & 0.224 & 0.381 \\
Unigram GPT-4o & 0.000 & 0.134 & 0.187 & 0.399 \\
Unigram Claude & 0.222 & 0.146 & 0.165 & 0.403 \\
Unigram RightAlign & 0.000 & 0.140 & 0.175 & 0.399 \\
SuperBPE GPT-4o (bal.) & 0.068 & 0.159 & 0.180 & 0.383 \\
PA-BPE NFC GPT-4o & 0.130 & 0.165 & 0.203 & 0.388 \\
MinGram SCRIPT-enc & 0.016 & 0.177 & 0.199 & 0.379 \\
BPE SCRIPT-enc GPT-4o & 0.136 & 0.171 & 0.205 & 0.378 \\
BPE NFC clean (bal.) & 0.218 & 0.159 & 0.177 & 0.381 \\
SuperBPE clean C2 (bal.) & 0.202 & 0.134 & 0.241 & 0.391 \\
\crnew{BPE NFC clean plus2 (bal.)}$^{e}$ & 0.114 & 0.152 & 0.212 & 0.379 \\
\crnew{BPE NFC clean plus3 (bal.)}$^{e}$ & 0.220 & 0.171 & 0.197 & 0.383 \\
\crnew{SuperBPE clean C3 (bal.)}$^{e}$ & 0.190 & 0.140 & 0.171 & 0.390 \\
\crnew{MinGram SCRIPT-enc (nl-split)}$^{e}$ & 0.216 & 0.152 & 0.235 & 0.383 \\
\midrule
Mistral-Nemo$^\dagger$ & 0.000 & 0.152 & 0.184 & 0.378 \\
LLaMA-3$^\dagger$ & 0.082 & 0.171 & 0.233 & 0.378 \\
\bottomrule
\end{tabular}
 \end{table*}

\cref{tab:mc20b-results} reports the math+code models' generation benchmarks and code BPB. Under identical training data, architecture, and budget, MBPP pass@1 ranges from $0.000$ to $0.250$ across the \crnew{\nMCExtended{} custom rows}, while HumanEval pass@1 ranges from $0.134$ to $0.177$. The paragraphs below examine how the code-structure metrics of \cref{sec:tokeval} relate to that range.

\paragraph{Code-structure metrics in the math+code models.}
One intrinsic property of a vocabulary is used as a control variable throughout this
paragraph: the number of vocabulary tokens that contain a line break (i.e., that contain a literal \colorbox{gray!20}{\texttt{\textbackslash n}}). The companion code blog post to this paper\crnew{\footnote{\url{https://cimeister.github.io/blog/code_tokenizer_ablations}}} reports how that count relates to MBPP pass@1 across a larger set of
tokenizers and models, together with the pretokenizer clause that determines
it.\footnote{On the math+code panel: across the \nMCByteLevel{} custom models with
byte-level vocabularies, for which token-byte counts are comparable, the count of vocabulary
tokens containing a line break correlates with MBPP pass@1 at Spearman $\rho = -0.717$
($p = 8.1\times10^{-4}$), and MBPP pass@1 averages $0.075$ for models whose tokenizer has at
least one token fusing punctuation with a following line break, against $0.226$ for models
whose tokenizer has none (Mann--Whitney $p = 4.5\times10^{-4}$).}
In the math+code models, AST boundary alignment correlates with MBPP pass@1 at $\rho = +0.\crnew{65}$
(significant, $p_{\text{adj}} = 0.036$, $n = \nMCPrimary{}$). The adjustment is the
Benjamini-Hochberg procedure (described in \cref{app:stat-protocol}) applied within one family of 38 tests. This is the only
significant code-structure correlation we find in our setup, which uses the \nMCPrimary{} custom tokenizers of \cref{tab:mc20b-results} (reference tokenizers are not included).
We do not interpret this correlation as an effect of AST boundary alignment itself, for two reasons. First, this association is not separable from the line-break
handling described in the companion post. Controlling for the number of vocabulary tokens containing a line
break reduces the measured association, giving a partial correlation\crnew{\footnote{A partial rank
correlation is the correlation between two quantities after the part of each that is predicted
by a third quantity has been removed from both.}} of $\rho = +0.\crnew{26}$ (not significant\crnew{, $p=0.31$}, $n = \nMCByteLevel{}$\crnew{, this panel's members with byte-level vocabularies: the two SCRIPT-encoding tokenizers are excluded because their vocabularies are built from script-block units, so token-byte counts are not comparable}). This is consistent with the check in \cref{sec:results} that restricts the panel to tokenizers sharing one pretokenizer choice, where the same correlation reverses sign.
Second, the two MinGram tokenizers differ only in whether merges may cross line breaks (\cref{tab:tokenizer-grid}). MBPP pass@1 is $0.016$ for the MinGram tokenizer whose merges may cross them and $0.216$ for the MinGram tokenizer that splits there, a difference of $2.3$ panel standard deviations, where a panel standard deviation is the standard deviation of a measurement across this panel's tokenizers. Their respective AST
boundary alignments, on the other hand, barely differ; they round to the same value \cref{tab:intrinsic-flores}, differing by $0.0014$ of that metric's standard deviation across the panel's tokenizers. This is too
small a difference to account for the $0.200$ difference in MBPP pass@1. The splitting variant is printed in \cref{tab:mc20b-results} as MinGram SCRIPT-enc (nl-split), one of the rows marked $^{e}$. 
When considering the extended panel, which contains the \nMCExtended{} custom tokenizers of the extended math+code roster and enters every family variant rather than one representative per family, the same association is $\rho = +0.66$ (significant, $p_{\text{adj}} = 0.017$, $n=\nMCExtended{}$), matching the primary panel above in direction and in significance.

\paragraph{Operator isolation.}
Operator isolation, which measures whether operators are dedicated tokens rather than tokens fused with adjacent operands (\cref{sec:tokeval}), is computed separately on the prose, math, and code corpora. Its correlation with the math+code models' MBPP pass@1 is positive in all three measurement domains: $\rho = 0.52$ on prose, $0.34$ on code, and $0.28$ on math, with $n = \nMCPrimary{}$ custom tokenizers. None of the three correlations reaches significance after Benjamini-Hochberg adjustment, and the smallest adjusted $p$ is $0.058$, on the prose corpus. The panel contains two UnigramLM tokenizers for which MBPP pass@1 is exactly $0.000$, and removing those two brings all three correlations to adjusted significance ($p_{\text{adj}} = 0.012$, $0.007$, and $0.007$, with $n=18$). 
Because significance depends on those two models, we read the association as suggestive rather than established. 
Further, because the tokenizers that differ in operator isolation also differ in their whole pretokenization regex, we cannot safely attribute the association to operator handling, specifically, rather than some other aspect of the pretokenizers' handling of text.

\subsection{Per-Example Tokenization Metrics}\label{app:external-per-example}

The correlations above compare one model with another using an aggregate score. A second question is whether the
same metrics have any association with outcomes inside a single model. Concretely, we ask: among models attempting
the same problem, is a model whose tokenizer segments that problem's text into well-aligned
pieces more likely to solve it? We ask this question over two sets of models: this paper's own math+code models, and a set of publicly released code models trained independently of this pipeline.

\paragraph{Design.} For every pair of a model and a problem, we compute nine properties (our intrinsic metrics) of the way that model's own tokenizer segments the problem's text, covering AST boundary alignment, identifier fragmentation (\cref{sec:tokeval} defines each), and subcategories of the two. These properties are each computed twice, once on the prompt and once
on the reference solution.  
The analysis here relates the intrinsic tokenizer properties to the following binary outcome: 
whether that model's generation for that particular problem passed the tests (evaluation protocol for MBPP and HumanEval are described in \cref{app:eval-harness}).

We model the relationship between each of these properties and problem success using a conditional logistic regression model. 
In simple terms, this fit allows us to compare models within a problem.
Each problem forms its own ``group,'' and only the differences among models on that one problem contribute to the parameter that quantifies the relationship between the tokenizer property of interest and problem success.\footnote{A problem that every model solves, or that no model solves, does not contribute to the conditional likelihood.} Any property of the problem itself, including its difficulty and any feature of its text, is removed from the comparison by using this setup. The fit also includes one indicator term per model, which accounts for that model's overall tendency to solve problems. For fits that pool the two tasks, it includes one such term per model per task, for the reason given below. This term can be thought of as a control for overall model ability. 
As a second specification we refit with a Bayesian generalized linear mixed model in which each problem has its own random intercept, which allows us to account for problem difficulty in a different way.

One regression model is fitted for each combination of a tokenizer code metric, a text basis (the prompt or the
reference solution), and a task scope (MBPP alone, HumanEval alone, or the two pooled).
$p$-values are adjusted with the Benjamini-Hochberg procedure within each family of tests, where each text basis and task scope combination forms a test family. 
The set of external models that we evaluate consists of 5 publicly released code models
(StarCoder2-3B, Qwen2.5-Coder-3B, CodeGemma-2B, phi-2, stable-code-3b), each with its own
tokenizer, spanning 5 distinct tokenizer classes and vocabularies from 49{,}152 to 256{,}000 tokens, evaluated on HumanEval (164 problems) and MBPP (500 problems). 
The internal set of models we consider consist of the \nMCPrimary{} custom math+code tokenizers of \cref{tab:mc20b-results}, with their own MBPP
and HumanEval generations.

\paragraph{\crnew{No association within either set of models.}}
\crnew{On the generations from our models, 53 regression models are fit, varying the specific tokenizer code metric, the two text bases, and the three task scopes. Nine metrics, two text bases, and three scopes give 54 combinations; the remaining one, keyword full-alignment measured on the MBPP reference solutions, takes the value $1.0$ on every MBPP row, so no fit exists for it. The estimate for the parameter quantifying the strength of the relationship between the code intrinsic metric and performance is not significant after adjustment in any of the fits. Among the fits that only consider one task, the smallest adjusted $p$ is $0.087$, for the identifier fragmentation rate measured on the
reference solution and evaluated on MBPP.}
Evaluating the external model set gives the same result. 

\paragraph{\crnew{Pooling the two benchmarks requires one ability term per benchmark.}}
MBPP and HumanEval have different overall pass rates, and the tokenization metrics also
take systematically different values on the two benchmarks' texts. 
Results when fitting a regression model that uses data from both tasks, using one ability term per benchmark, are in line with those from the per-task fits above.
Notably, a pooled fit that gives each model a single ability term shared across
both benchmarks cannot match both pass rates at once. 
The difference between the two benchmarks then contributes to the metric's coefficient instead, because the metric's values also differ between the two benchmarks. 
Ultimately, a spurious association can appear in a fit using a pooled term that is absent from each benchmark taken alone.
We directly tested the design choice of using the shared term using a likelihood-ratio test. The test compares the fit when one pooled term across tasks is used to the fit when an ability term per task is used, asking whether the single-term constraint makes the fit worse by more than
chance variation would explain. On the internal set, the constraint of one ability term per model
shared across both benchmarks is rejected in all 18 pooled tests, with test statistics from
$137.8$ to $193.5$ on $20$ degrees of freedom. We therefore report results for each benchmark
separately, and report no pooled-ability-term results as findings.

Taken together, the results in this appendix support reading the code-structure metrics as
audits of what a vocabulary contains rather than as predictors of a benchmark score. On this set of models their correlation with code generation is not separable from line-break handling, and
within a single model they show no association with which problems that model solves.

\end{document}